\documentclass[lettersize,journal,compsoc,article]{IEEEtran}
\usepackage{amsmath,amsfonts}
\usepackage{array}
\usepackage[caption=false,font=normalsize,labelfont=sf,textfont=sf]{subfig}
\usepackage{textcomp}
\usepackage{stfloats}
\usepackage{url}
\usepackage{verbatim}
\usepackage{graphicx}
\usepackage{cite}
\usepackage{orcidlink}
\usepackage{pifont}
\usepackage{lipsum}
\usepackage[hang,flushmargin]{footmisc}

\usepackage{bbding}
\usepackage{amsmath, bm}
\usepackage{graphicx}
\usepackage{bbold}
\usepackage{multirow}
\usepackage{threeparttable}
\usepackage{caption} 
\usepackage{balance}
\usepackage{multirow}
\usepackage{booktabs}
\usepackage{hyperref}
\usepackage{dsfont}
\usepackage{enumitem}
\usepackage{arydshln}
\usepackage{xcolor,colortbl}

\usepackage{cite}
\usepackage{hyperref} 
\usepackage{amsmath,amssymb,amsfonts}
\usepackage{algorithmic}
\usepackage{graphicx}
\usepackage{textcomp}
\usepackage{xcolor}
\def\BibTeX{{\rm B\kern-.05em{\sc i\kern-.025em b}\kern-.08em
    T\kern-.1667em\lower.7ex\hbox{E}\kern-.125emX}}

\usepackage{amsmath}
\usepackage{amsfonts}
\usepackage{amsthm}
\newtheorem{theorem}{Theorem}
\newtheorem{lemma}{Lemma}

\newtheorem{definition}{Definition}
\newtheorem{proposition}{Proposition}

\usepackage{wrapfig}
\usepackage{xcolor}         
\usepackage{multirow}
\usepackage{mathrsfs}
\usepackage{amsmath}
\usepackage{graphicx}
\usepackage{algorithm}
\usepackage{algorithmic}
\usepackage{tabularx}
\usepackage{makecell}
\usepackage{CJKutf8}
\usepackage{balance}
\usepackage{multirow}
\usepackage{booktabs} 
\usepackage{xcolor}
\usepackage{soul}
\usepackage{tabularx} 
\usepackage{array}
\usepackage{subcaption}
\usepackage{hhline}
\usepackage{graphicx} 
\usepackage{float}
\usepackage{amsmath} 
\usepackage{amsthm} 
\usepackage{orcidlink}
\definecolor{gray}{gray}{0.93}
\usepackage{arydshln}
\usepackage{tikz}

\definecolor{purple}{RGB}{112,48,160}
\definecolor{ocean}{RGB}{2,154,152}
\definecolor{blue}{RGB}{31,119,180}
\definecolor{red}{RGB}{192,0,0}

\newcommand{\NFFst}[1]{{{#1}}}
\newcommand{\Fst}[1]{{\textbf{\textcolor{blue}{#1}}}}
\newcommand{\XFst}[1]{{\textbf{\textcolor{red}{#1}}}}

\newcommand{\std}[1]{$_{#1}$}

\begin{document}

\title{Learnable Game-theoretic Policy Optimization for Data-centric Self-explanation Rationalization}

\author{\IEEEauthorblockN{
Yunxiao Zhao~\orcidlink{0000-0002-9133-7324}, 
Zhiqiang Wang~\orcidlink{0000-0002-9269-3988}, 
Xingtong Yu~\orcidlink{0000-0002-2884-8578},  
Xiaoli Li~\orcidlink{0000-0002-0762-6562}, 
Jiye Liang~\orcidlink{0000-0001-5887-9327}, 
and Ru Li~\orcidlink{0000-0003-1545-5553}, 
}\\
\thanks{
This manuscript is under review by IEEE. 
Yunxiao Zhao, Zhiqiang Wang, Jiye Liang, Ru Li are with School of Computer and Information Technology, Shanxi University, Taiyuan, China (e-mail: yunxiaomr@163.com, \{wangzq,ljy,liru\}@sxu.edu.cn). 
Xingtong Yu is with the School of Computer and Information Systems, Singapore Management University, Singapore (yxt95@mail.ustc.edu.cn).
Xiaoli Li is with the Institute for Infocomm Research, A*Star, Singapore (xlli@i2r.a-star.edu.sg).
This work has been submitted to the IEEE for possible publication. Copyright may be transferred without notice, after which this version may no longer be accessible. \\
}}


\markboth{}%
{Shell \MakeLowercase{\textit{et al.}}: A Sample Article Using IEEEtran.cls for IEEE Journals}


\IEEEtitleabstractindextext{%
\begin{abstract}
Rationalization, a data-centric framework, aims to build self-explanatory models to explain the prediction outcome by generating a subset of human-intelligible pieces of the input data. 
It involves a cooperative game model where a generator generates the most human-intelligible parts of the input (i.e., rationales), followed by a predictor that makes predictions based on these generated rationales.
Conventional rationalization methods typically impose constraints via regularization terms to calibrate or penalize undesired generation. However, these methods are suffering from a problem called mode collapse, in which the predictor produces correct predictions yet the generator consistently outputs rationales with collapsed patterns. 
Moreover, existing studies are typically designed separately for specific collapsed patterns, lacking a unified consideration.
In this paper, we systematically revisit cooperative rationalization from a novel game-theoretic perspective and identify the fundamental cause of this problem: the generator no longer tends to explore new strategies to uncover informative rationales, ultimately leading the system to converge to a suboptimal game equilibrium (correct predictions \textit{v.s} collapsed rationales).
To solve this problem, we then propose a novel approach, Game-theoretic \textbf{P}olicy \textbf{O}ptimization oriented \textbf{RAT}ionalization (\textsc{PoRat}), which progressively introduces policy interventions to address the game equilibrium in the cooperative game process, thereby guiding the model toward a more optimal solution state.
We theoretically analyse the cause of such a suboptimal equilibrium and prove the feasibility of the proposed method. 
Furthermore, we validate our method on nine widely used real-world datasets and two synthetic settings, where \textsc{PoRat} achieves up to 8.1\% performance improvements over existing state-of-the-art methods. 
\end{abstract}

}

\maketitle

\section{Introduction}\label{introduction}

\IEEEPARstart{W}ith the success of deep learning in processing large-scale data, the demand for model interpretability has garnered significant attention in recent years \cite{lipton2018mythos}. 
Ideally, model explanations should be both plausible and faithful, which means they should be aligned with human understanding and can reflect the model's predictive behaviour simultaneously \cite{jacovi-goldberg-2020-towards,chan2022unirex}. 
Early studies of explainability \cite{bhattacharya2022applied,ribeiro2016should,lundberg2017unified,ribeiro2018anchors,menon-etal-2023-mantle}, focusing on model-centric explanations, try to leverage post-hoc explanation by approximating important features through machine learning models to explain predictions. Despite appearing plausible, this may not faithfully represent an agent’s decision \cite{lipton2018mythos}, since the explanation generation process is trained separately from the model predictions. 
In contrast to model-centric post-hoc methods, data-centric self-explanation techniques typically offer increased transparency and faithfulness \cite{yu20193player}, as the prediction is made based on the explanation itself \cite{liu2025exploring,liu2025adversarial}.

\begin{figure}[t]
    \centering
    \includegraphics[width=0.95\columnwidth]{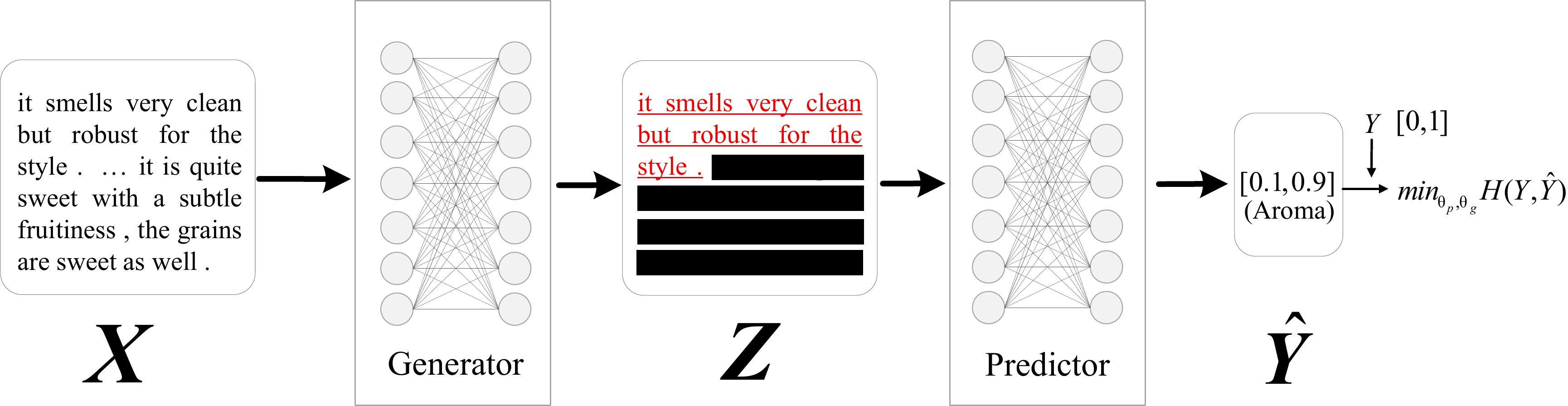}  
    \caption{A standard RNP framework on the binary sentiment analysis, where $X$, $Z$, $\hat{Y}$, $Y$ represent the input data, rationale, prediction and the groundtruth label, respectively. } 
    \label{fig:RNP}
    \vspace{-10pt}
\end{figure}

In this study, our primary focus is on investigating a general data-centric self-explaining framework called Rationalizing Neural Predictions (RNP, also known as rationalization) \cite{lei2016rnp}, which with its variants has become mainstream approach for facilitating the interpretability of models \cite{yuetowards,zhao-etal-2024-agr,liu2024enhancing,g-rat,jiang-etal-2024-mare,2023cr,storek-etal-2023-unsupervised,liu2023decoupled,liu2022fr}, and also has the potential to be applied to downstream tasks such as sentiment analysis \cite{lei2016rnp,sha2023rationalizing}, image classification \cite{yuan2020interpreting}, graph neural networks \cite{luo2020parameterized}, legal judgment \cite{yue2024circumstance}, and the recommendation system \cite{Deng2023MultiAspectIN}. 
As illustrated in Fig.\ref{fig:RNP}, there is a standard rationalization RNP framework, which aims to generate a small and human-intelligible subset (i.e., rationale) from the input data to support and explain the prediction results when yielding them.
Here, they highlight key textual spans for input data and utilize a cooperative game with two players (a generator and a predictor) to maximize prediction accuracy through the computation of the maximum mutual information (MMI) loss \cite{yu2021A2R}. 
As a result, this principle aims to faithfully provide explanations to explain the coupled connection between the input and the model-agnostic task label \cite{liu2024mrd}.

\begin{figure}[t]
    \centering
    \includegraphics[width=0.99\columnwidth]{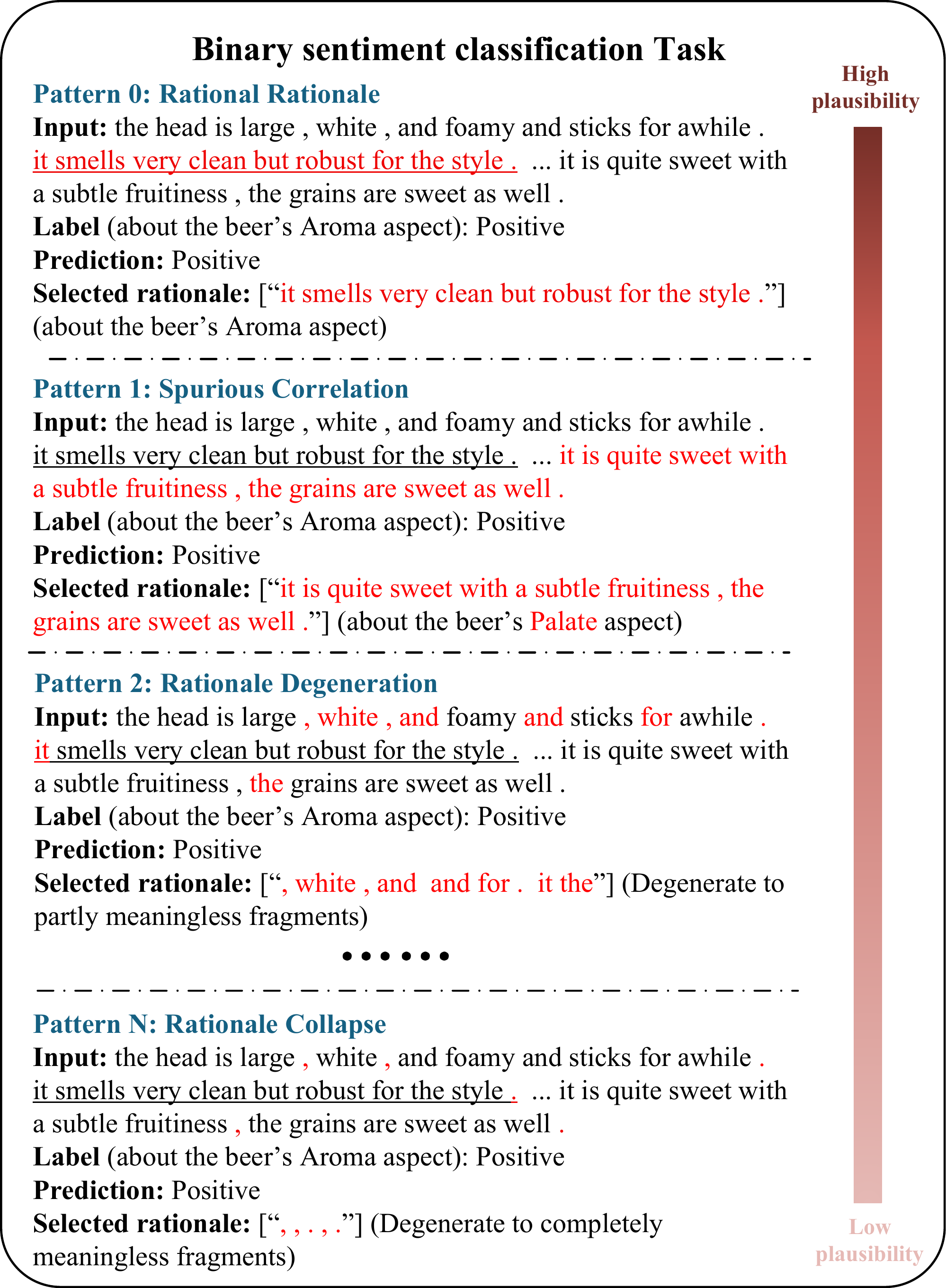}  
    \caption{Rationales of different patterns illustrate the rationales of rational, spurious correlation, partial degeneration, and complete degeneration, which are caused by the generator. Human-annotated rationales are underlined; “\textcolor{red}{red} text” indicates rationales generated by models. }  
    \label{fig:2}
    \vspace{-14pt}
\end{figure}

Despite such a rationalization model can ensure the faithfulness of the model \cite{yu2021A2R} (i.e., certification of exclusion \cite{liu-etal-2023-mgr}),
the cooperative game is difficult to train if the generator and the predictor are not well coordinated. In this paper, we identify two key challenges that constrain the learning and optimization of the rationale within this self-explaining framework.
\textbf{i) Mode collapse of rationale generation. }
Mode collapse refers to the phenomenon where, during the process of generating self-explanations, the predictor produces correct predictions, yet the generator consistently outputs collapsed rationale patterns. It becomes fixated on a few dominant modes in the training data and fails to capture the rational rationale distribution.
As illustrated in Fig.\ref{fig:2}, $N$ different rationale modes prevent the generator from generating meaningful rationales with high plausibility. 
The generator may produce some meaningless fragments that are decorrelated information (e.g., Pattern 1) or not human-intelligible (e.g., Pattern N)  to explain the predictor's prediction on the Aroma aspect. 
Though the predictor infers correct predictions, the generator yields uninformative rationales, converging to a sub-optimal state (correct predictions \textit{v.s} collapsed rationales).
The core idea lies in the fact that the generator initially produces a specific pattern (maybe bad patterns), when passed to the predictor, which still leads to a correct label. In such cases, the generator can receive positive feedback and is thus encouraged to overfit to that particular pattern. 
\textbf{ii) Unified modeling of rationale patterns. }
Most existing research focuses on individual cases, lacking of unified modeling for data-intrinsic rationale patterns.  
For example, a series of studies adopts causal inference and a rectified criterion to exclude the mode collapse of spurious correlations (Pattern 1).
Chang et al. \cite{chang2020invariant} use an environment-agnostic predictor to recognize spurious correlations;  
Yue et al. \cite{yue-etal-2023-interventional} aim to remove spurious correlations based on backdoor adjustment; 
Liu et al. \cite{liu2023mcd} propose the minimum conditional dependence criterion to uncover causal rationales rather than spurious features. 
In addition, some studies use additional information to regularize the predictor to address the partial degeneration problem (Pattern 2). 
Yu et al. \cite{yu2021A2R} uses soft attention from the generator to input full text information into the predictor; 
Huang et al. \cite{Huang2021DMR} and Liu et al. \cite{liu2024enhancing} follow the importance of full input and align from different points of view; 
Liu et al.  \cite{liu2022fr} uses the same encoder between the generator and the predictor to transmit information.
Although the above studies have made progress on these two patterns, the patterns present in the data are not finite in variety, as evidenced by recently identified rationalization failure \cite{g-rat}. Moreover, some studies \cite{liu2023mcd, liu2024mrd} also indicate that various data features may compete with the true rationale for extraction opportunities, thereby hindering the interpretability of the data.
This can easily lead to the generator developing diverse rationales selected without human understanding.

To address the above problems, we unify spurious correlations, degeneration and other collapsed rationale fragments, treating them collectively as suboptimal rationales; and  systematically revisit the cooperative mechanism of rationalization from a novel game-theoretic perspective and present the existing cooperative problem in the rationalization framework. %
We theoretically analyze that the fundamental cause of collapsed rationales is that the generator no longer tends to explore new strategies and falls into a suboptimal game equilibrium. 
Therefore, to solve this problem, we propose a novel rationalization method and prove its feasibility from a game-theoretic perspective, termed \textbf{P}olicy \textbf{O}ptimization oriented \textbf{RAT}ionalization (\textbf{PORAT}), which aims to guide the rationalization model to cope with such a suboptimal equilibrium and to promote the mode collapse problem of cooperative rationalization. 
The contributions of this paper can be summarized as follows:
\begin{itemize}
\setlength{\itemsep}{0pt}
\item {New perspective:} We unify the concept of collapsed rationales, systematically revisit the cooperative game mechanism of rationalization from a novel perspective, and reveal the game-theoretic problem between two players, i.e., sub-optimal game equilibrium.



\item {Theoretical insights:} We theoretically analyze the fundamental causes of the sub-optimal game equilibrium problem between two players in rationalization.

\item {New methodology:} We propose a novel method called PORAT, which progressively introduces policy interventions to address the sub-optimal game equilibrium problem in the cooperative game process. Moreover, we prove the feasibility of the proposed PORAT.

\item {Empirical results:} Extensive experiments on real-world benchmarks (nine widely used datasets) and synthetic settings (two synthetic settings) demonstrate the effectiveness of PORAT, which improves the F1 score by up to 8.1\% as compared to the state-of-the-art method.

\end{itemize}

The remainder of this paper is organized as follows. Sect. \ref{related_work} summarizes the related works. The problem definition of rationalization is given in Sect. \ref{problem_definition}. The revisiting of cooperative rationalization is specified in Sect. \ref{motivation}, including game-theoretic mechanisms and problems.
The proposed method and theoretical insights are presented in Sect. \ref{method} 
Besides, the experimental results and analysis are in Sect. \ref{experiments}. Finally, we conclude this study in Sect. \ref{conclusion}. 
To streamline the main text for better readability, we have moved some non-critical technical proofs and setups to the Appendices.

\section{Related Work}\label{related_work}

Explainability is a critical research area in the fields of data science and artificial intelligence. In this section, we categorize the explainability research into three main types: data-centric explanations, model-centric explanations and generative explanations with large language models. 
our primary focus will be on the methods of rationalization in the domain of data-centric explanation research.

\vspace{-2pt}
\subsection{Data-centric rationalization explanations}
\vspace{-2pt}
Data-centric rationalization has received increasing attention in recent years, aiming to answer the question: Which parts of the input data drive the prediction made by deep neural networks (DNNs)? \cite{yuetowards}. Typically, this research consists of a generator and a predictor, which produces task-specific predictions using the predictor, while the generator identifies a short and coherent subset of the original input (namely rationale) that is sufficient to explain and support the prediction.
There are two main lines of research: {supervised rationalization} and {unsupervised rationalization}.

\noindent \textbf{Supervised rationalization}. 
The supervised rationalization framework jointly utilizes rationales and class labels during training.
Representative works mainly focus on benchmarks and proposed methods \cite{deyoung2020eraser, lehman-etal-2019-inferring,chan2022unirex,li2022unifying}. 
For example, 
DeYoung et al. \cite{deyoung2020eraser} propose an ERASER benchmark which contains several datasets with both task labels and gold rationales.
Lehman et al. \cite{lehman-etal-2019-inferring} propose a pipeline approach as a supervised baseline. 
Chan et al. \cite{chan2022unirex} develop a unified framework to replace previous works’
heuristic design choices with a generic learned rationale generator.
Li et al. \cite{li2022unifying} propose to employ mixed adversarial training and boundary match constraint to improve supervised rationales. 
These studies usually rely on gold rationales annotated by humans during model training, and formulate rationalization as a multitask learning problem, optimizing the joint likelihood of both class labels and extractive rationales.
However, for most tasks, obtaining large-scale annotated rationales is impractical, which limits the applicability in real-world scenarios. 

\noindent \textbf{Unsupervised rationalization.} The other major line of research is initiated by Lei et al. \cite{lei2016rnp}, who propose a unsupervised framework for self-explainable rationale learning. This approach also employs a generator and a predictor component. Since the predictor makes its decision solely based on the explanation produced by the generator, the resulting rationale is faithful to the model’s behavior \cite{lyu-etal-2024-towards, liu-etal-2023-mgr, yu2021A2R}.
However, optimizing unsupervised rationales remains a challenging problem. 
Some early studies \cite{bao-etal-2018-deriving,bastings2019HardKuma,paranjape2020ib} mainly focus on how to mitigate this blocking problem from a rationale sample perspective. 
Most recent studies have focused separately on individual collapsed problems. 

For example, a series of studies uses additional information alignment to regularize the predictor, aiming to directly improve the degeneration.
Yu et al. \cite{yu20193player} add a complementary predictor that uses text not selected as the rationale, and use soft attention from the generator to input full text information into the predictor \cite{yu2021A2R}.
Huang et al. \cite{Huang2021DMR} and Liu et al. \cite{liu2024enhancing} follow the importance of full input and align from different points of view.
Yue et al. \cite{yue2022dare} improve rational representations by reducing the mutual information between rational and non-rational parts of the input.
Liu et al. \cite{liu2022fr} use the same encoder between the generator and the predictor; they also introduce lipschitz continuity to model asymmetric learning rates, with the aim of decoupling the optimization frequency of two players \cite{liu2023decoupled}.
Zhao et al. \cite{zhao-etal-2024-agr} and Hu et al. \cite{g-rat} introduce an extra reinforced causal agent and a guidance module to guide the generator regulate the degeneration process of rationalization, respectively.

On the other hand, some work introduces causal theory and a rectified criterion to address the problem of spurious correlations.
Chang et al. \cite{chang2020invariant} use an environment-agnostic predictor to recognize spurious correlations.
Yue et al. \cite{yue-etal-2023-interventional} aim to remove spurious correlations based on backdoor adjustment.
Liu et al. \cite{liu2023mcd} propose the minimum conditional dependence criterion to uncover causal rationales rather than spurious features, also introduce a new criterion that treats spurious features as equivalent to plain noise.

Although the above methods achieve improvements on individual collapsed mode problems, few studies have investigated the underlying nature of these collapsed rationales and conducted unified modeling. Moreover, the interaction between the two players in rationalization has not been fully explored.
In this paper, we analyze the coordination mechanism of rationalization using game-theoretic methodology from a novel perspective, aiming to systematically reveal relationships and underlying problem between two players, and propose a solution to address this problem.

\vspace{-2pt}
\subsection{Model-centric post-hoc explanations}
\vspace{-2pt}
Different from data-centric methods, model-centric methods aim to approximate the important features used by machine learning models to generate predictions, which has also been widely explored \cite{bhattacharya2022applied}, such as LIME \cite{ribeiro2016should}, SHAP \cite{lundberg2017unified}, Anchors \cite{ribeiro2018anchors} and so on. 
Recently, Menon et al. \cite{menon-etal-2023-mantle} propose MaNtLE, a natural language explainer, to analyze a set of classifier predictions and generate natural language explanations for structured classification tasks. 
From the perspective of explanation provenance, these methods are commonly known as post-hoc explanation approaches, as the explanations they provide are generated independently of the well-trained predictor. As a result, post-hoc explanation usually provides less transparency \cite{lipton2018mythos} and faithfulness \cite{yu20193player}. 
Therefore, these explanation methods are a research line that is related but orthogonal to our research.

\vspace{-2pt}
\subsection{Generative explanation with large language models}
\vspace{-2pt}
With the great success of large language models, a new line of research in explainability has emerged: in-context learning (ICL)-based chain-of-thought (CoT) reasoning \cite{wei2022chain}. Instead of identifying rationales from the input data, these methods generate intermediate reasoning steps before producing an answer, treating the reasoning process itself as an explanation. This compelling CoT technique has inspired several variants such as IRCoT \cite{ircot}, Self-ASK \cite{press-etal-2023-measuring}, FLARE \cite{jiang-etal-2023-active} and DRAGIN \cite{su-etal-2024-dragin}, all of which have shown promising progress. 
However, due to unpredictable failure problems \cite{kiciman2023causal} and hallucinated reasoning \cite{ji2023survey}, CoT-based generative explanations produced by large language models are often unreliable in high-stakes scenarios. Recent studies suggest that language models still struggle with unsupervised, self-explanatory tasks \cite{2023cr,liu2023mcd}, also CoT-based language model reasoning is not always faithful \cite{arcuschin2025chain,turpin2023language}.

\section{Problem Definition}\label{problem_definition}

\noindent \textbf{Notations.} 
In this study, without losing generality, we consider the classification problem and denote the generator and predictor as $f_{G}$(·) and $f_{P}$(·), with $\theta_{g}$ and $\theta_{p}$ representing their parameters. 
Here, to ensure clarity and facilitate better comparison with mainstream methods, we consider input $X$ as text data. Therefore, the input can be represented as $X = [x_1,x_2,...,x_l] (1 \leq i \leq l)$ consisting of text tokens $x_i$, where $l$ is the number of tokens. The label of $X$ is a one-hot vector $Y \in \{0,1\}^{c}$, where $c$ is the number of categories.

\noindent \textbf{Self-explanation rationalization.}
The standard rationalization framework RNP consists of a generator $f_{G}$(·) and a predictor $f_{P}$(·), where the generator aims to select the most informative pieces from the input $X$. 
For each $(X,Y) \in \mathcal{D}$, the generator first gets a sequence of binary masks $M = [m_1,m_2,...,m_k] \in \{0,1\}^l$. Then, it forms the rationale $\hat Z$ by the element-wise product of $X$ and the binary mask $M$: 
\begin{equation}\label{Eq1}
     \hat Z = M \odot X = [m_1x_1,m_2x_2,...,m_kx_k].
\end{equation}
Subsequently, the informativeness of the rationale $\hat Z$ generated by $f_{G}$(·) is measured by the negative cross entropy $-H(Y,Y_{\hat z})$, 
where $Y_{\hat z}$ is the output of $f_{P}$(·) with the input being $\hat Z$. 
Consequently, the generator and the predictor are usually optimized cooperatively:
\begin{equation}\label{Eq2}
     \min_{\theta_G,\theta_P} \mathcal{H}(Y,\hat{Y}\mid f_G(X)), s.t. \hat{Y}=f_P(f_G(X)). 
\end{equation}
Here, we denote the rationale $\hat Z$ generated by $f_G(X)$, i.e., $\hat Z=f_G(X)$. 
Ideally, the rationale $\hat Z$ by the model should consist of meaningful rationales with best plausibility, which we denote best rationale as $Z$ (named golden rationale\footnote{Note that the golden rationale called refers only to the rationale under the assumed ideal conditions, the actual task is unsupervised rationalization during model training \cite{yuan2025boosting,jiang-etal-2024-mare}.}).

\noindent \textbf{Sparsity and continuity constraints.} 
To make the rationales generated by $f_{G}$(·) human-intelligible, RNP methods usually constrain the rationales by compact and coherent regularization terms. 
Thus we also adopt the same constraints used in most research:
\begin{equation}\label{Eq3} 
	\Omega(M) = \lambda_1 \left\lvert \frac{{\lvert\lvert M \rvert\rvert}_1} {l}  - s \right\rvert + \lambda_2 \sum_{t} \lvert m_t - m_{t-1} \rvert
\end{equation} 
where 
$l$ denotes the number of tokens in the input. The first term encourages that the percentage of the tokens being selected as rationales is close to a pre-defined level $s$. The second term encourages the rationales to be coherent.
Finally, the overall objective learned is defined as 
\begin{equation}\label{Eq4}
	\min \mathbb{L} = \min_{\theta_G,\theta_P}\mathcal{H}(Y,\hat{Y})+\Omega(M),s.t.\hat{Y}=f_P(f_G(X)). 
\end{equation}

\section{Revisiting Cooperative Rationalization}\label{motivation}  

In this section, we first present the knowledge of preliminaries (Sect. \ref{4.0}) to illustrate the coordination mechanism of the RNP models (Sect. \ref{4.1}). 
Then, we analyze relationships and the underlying problem between two players from a game-theoretic perspective (Sect. \ref{4.2}).

\begin{figure}[t] 
    \centering
    \includegraphics[width=0.95\columnwidth]{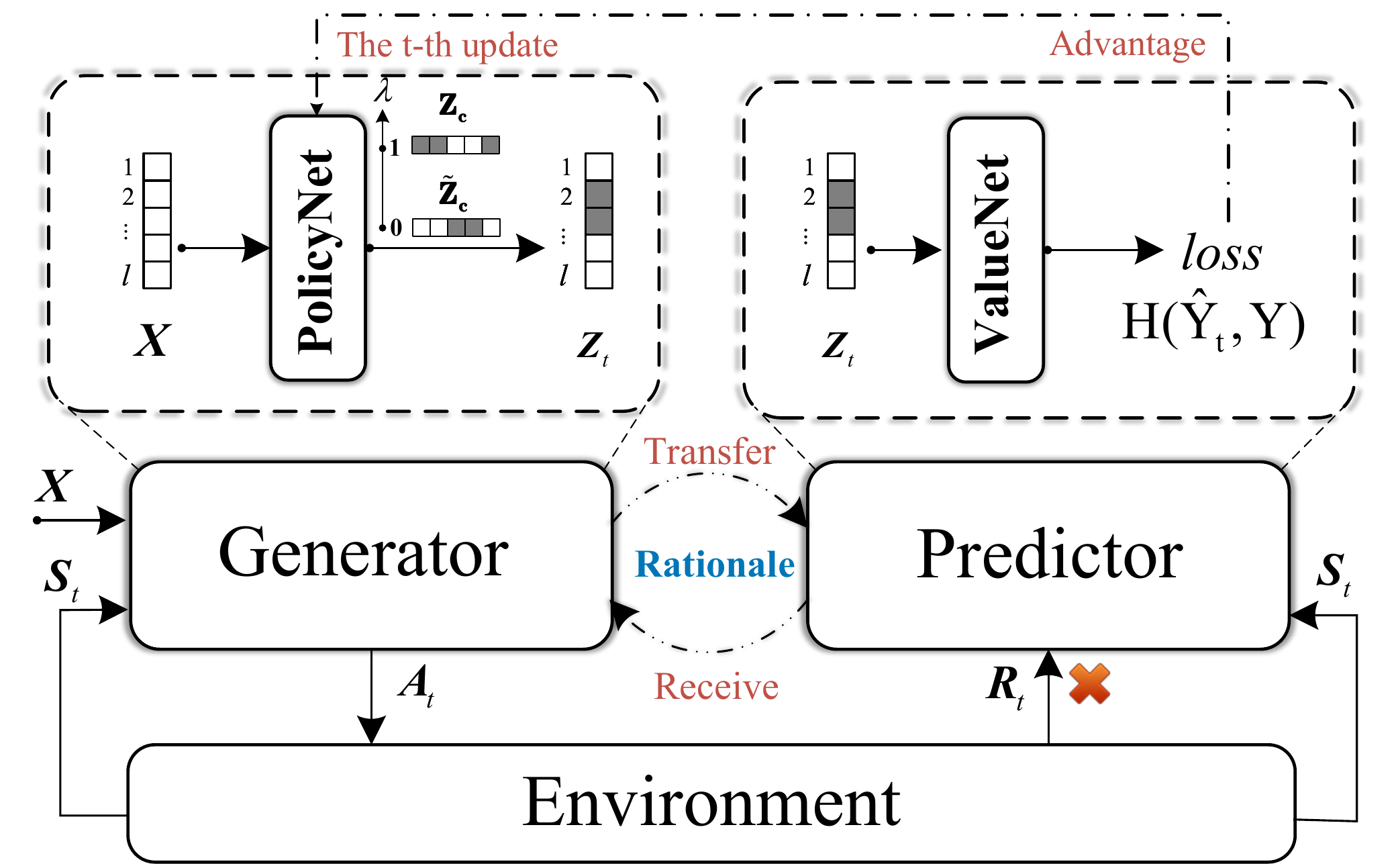}  
    \caption{The coordination mechanism in RNP framework.} 
    \label{fig:Intro}
    \vspace{-8pt}
\end{figure}

\subsection{Preliminaries}\label{4.0}

To intuitively illustrate the game-theoretic dynamics of RNP models, we first define the rationale optimization process, following prior work. 
We model the training process of rationale as a Markov decision process $\mathcal{M} = \{ \mathcal{S},\mathcal{A},\mathcal{P},\mathcal{R} \}$ \textit{from the generator perspective} \cite{zhao-etal-2024-agr}, where $\mathcal{S}=\{s^t\}$ represents set of states abstracting the process of optimizing rationale during training,
and $\mathcal{A}=\{a^t\}$ indicate the set of actions that update a rationale to the one state.
In particular, the transition dynamics $\mathcal{P}(s^{t+1}|s^t,a^{t+1})$ specify how the state $s^{t+1}$ is updated from the prior state $s^{t}$ by taking action $a^{t+1}$. 
Besides, $\mathcal{R}(s^{t},a^{t+1})$ quantifies the reward obtained after taking action $a^{t+1}$ based on the prior state $s^{t}$. 
Therefore, cooperative training for rationale can be depicted as the sequence process $(s^0,a^1,r^1,s^1,...,a^K,r^K,s^K)$,
where the state $s^t$ at timestep $t$ can be formulated by $s^t = \hat Z_t$ in the \textit{t-th} update.
However, previous work \cite{yu2021A2R,zhao-etal-2024-agr} neglects the involvement of the predictor. To this end, we introduce the following definitions, which enable us to derive game-theoretic policies of both the generator and the predictor during the game. 

\begin{definition}[Two-Agent Markov Games for Rationalization]\label{def:MAMGs}
Let $\mathcal{M}$ be a markov game process with two agents. It can be defined as a 7-tuple $<\mathcal{N}, \mathcal{S},\mathcal{A},\mathcal{P},\boldsymbol {\rho}_0,\gamma,\mathcal{R}>$ of states $\mathcal{S}$, actions $\mathcal{A}$, transition probability function $\mathcal{P}(s^{t+1} \mid s^t,a)$, the initial state distribution ${\rho}_0$, a discount factor $\gamma$, and the joint reward function $\mathcal{R}$, where $\mathcal{N}=\{1,2\}$.
\end{definition} 
\begin{definition}[Game-theoretic Policy for Rationalization]\label{def:MS}
A game-theoretic policy $\pi(a\mid s^t)$is a probability distribution defined as $\pi:{\mathcal{S}\times\mathcal{A}}\mapsto[0,1]$ indicating the probability of choosing
an action given the state $s$ at timestep $t$. In this paper, we follow previous work \cite{yu2021A2R,zhao-etal-2024-agr} and target on generate-predict strategies to learn a conditional policy $\pi(a\mid s^t)$, which indicates a policy for the generation or prediction of a candidate rationale.
\end{definition}

\subsection{Game-theoretic Mechanism for Rationalization}\label{4.1}

As shown in Fig.\ref{fig:Intro}, a standard RNP game consists of a generator $f_G$, a predictor $f_P$ and an environment $\mathbb{E}$. 
The $f_G$ produces a rationale and feeds this action back to the environment.
Meanwhile, $f_P$ receives a rationale and performs a fitting, guiding $f_G$'s policy update. 
Furthermore, $\mathbb{E}$ provides the corresponding observations and rewards based on the actions of $f_G$, respectively. 
However, it lacks a direct reward for $f_P$ since model faithfulness is a key concern. $f_P$ can only receive those rationales from $f_G$ \cite{liu-etal-2023-mgr,yu2021A2R,liu2023decoupled}. 
This leads to a key characteristic: \textit{the quality of rationales generated by the $f_G$ relies on the $f_P$'s supervision; and whatever the $f_G$ transmits, the $f_P$ receives}. 
Therefore, we can derive a proposition for the nature of rationalization as follows,
\begin{proposition}[Dependence and Non-discriminatory]\label{def:prop}
Given an RNP model, which consists of a generator $f_G$ and a predictor $f_P$.
Let $X$, $Y$ and $Z$ be the input data, label and rationale, where $\hat Z_i$ and $\hat Z_j$ are two candidate rationales. Then we have 
\begin{itemize} 
\item Dependence (for Generator): $\min_{\theta_G}\mathcal{H}(Y,\hat Y),s.t.\hat Y=f_P(\hat Z);\hat Z=f_G(X)$, which means 
learnable parameter $\theta_G$ depends on the $f_P$'s supervised loss.
\item Non-discriminatory (for Predictor): $\hat Y=f_P(\hat Z_i)$ and $\hat Z_i = Z$, which is satisfied for $f_P$; however, $\hat Y=f_P(\hat Z_j)$ and $\hat Z_j \neq Z$, which is also satisfied for $f_P$. 
\end{itemize}
\end{proposition} 
\noindent This implies that to ensure interpretability (faithfulness), the rationalization model indirectly compromises the reward optimization process inherent in the game. We further investigate this mechanism by providing insights from both \textit{probabilistic} and \textit{suboptimality} perspectives.

Ideally, the rationale candidate generated by the generator at time $t$ should be the most informative text segment, while simultaneously constrained by Equation \ref{Eq3}, if and only if there is exactly one, i.e., the gold rationale (we denote that $Z_c$).
In this context, the generator's policy network produces an action profile based on its distribution, thereby obtaining a candidate rationale $Z_t$ as the current state $s_t$.
Considering probabilistic events, the goal is to learn a high-quality of rationale $Z_c$ from $l$ masked tokens. The total number of possible events in the sample space is $2^l$, while the probability of the model learning $Z_c$ is only $\frac{1}{2^{l}}$. 
\textit{Therefore, there is a low probability of identifying a high-quality rationale in an unsupervised setting.}
On the other hand, $Z_c$ at time $t$ contains the higher informative piece, and the failed rationale $\tilde{Z}_c$ ($Z_c$'s complementary set for $X$), which contains the lowest informative one. 
Generally, for the MMI loss, we have $L_{MMI} (Z_c) << L_{MMI}(\tilde{Z}_c)$ \cite{liu2024mrd}. When the generator samples a rationale candidate $Z_t$, the loss can be expressed as
\begin{equation}\label{Eq110_} 
	L_{MMI} (Z_t)=\lambda L_{MMI} (Z_c)+(1-\lambda)L_{MMI}(\tilde{Z}_c),
\end{equation}
where $\lambda=d(Z_t,Z_{c})$ represents the distance between the generated candidate rationale $Z_t$ and the gold rationale $Z_c$ \cite{liu2023decoupled}.
Based on the above, we derive the following lemma.
\begin{lemma} \label{lem:sub}
Let $L_{MMI} (Z_c)$ and $L_{MMI}(\tilde{Z}_c)$ represent the MMI loss corresponding to the gold rationale and the failed rationale, respectively. 
Then, existing at least one suboptimal state $Z_t$ such that, 
\begin{equation}\label{Eq} 
	 L_{MMI}(Z_c)<L_{MMI} (Z_t)<L_{MMI}(\tilde{Z}_c). 
\end{equation}
\end{lemma} 
\noindent When $\lambda=1$ holds, $L_{MMI}(Z_t)=L_{MMI}(Z_c)$ that is relatively small; 
when $\lambda=0$ holds, $L_{MMI}(Z_t)=L_{MMI}(\tilde{Z}_c)$ that is relatively large; and when $0<\lambda<1$ holds, the loss of $Z_t$ falls between the two. 

In summary, within the game-theoretic framework for rationalization, unsupervised optimization rarely leads directly to a high-quality rationale. 
\textit{Once a low-quality rationale is selected, the model is likely to converge to a suboptimal state.}

\begin{figure*}[t]
    \centering
    \includegraphics[width=2.00\columnwidth]{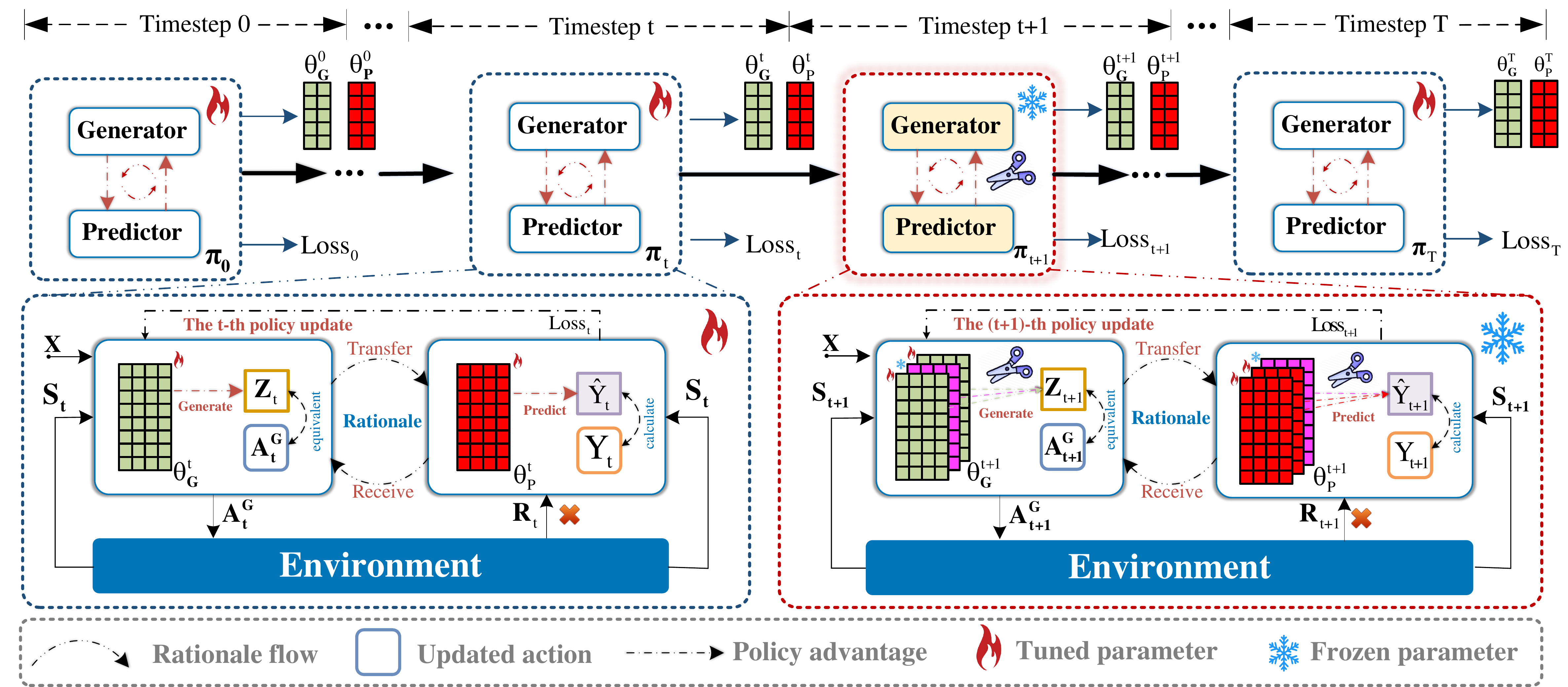}  
    \caption{The overview architecture of PORAT, where $X,Z,\hat{Y},Y$ represent the input text, rationale, prediction and the groundtruth label, respectively. Here, we provide policy intervention at $(t+1)$-th timestep, which is a progressive policy optimization to help the whole model escape a suboptimal state.} 
    \label{fig:PORAT}
\end{figure*}

\subsection{Game-theoretic Problem for Rationalization}\label{4.2}

Then, what leads to such a sub-optimal state? 
To analyze the cooperative correlation between the generator $f_G$ and the predictor $f_P$, we use the actor-critic-based process
\cite{sutton2018reinforcement,degris2012model,pmlr-v48-mniha16} to present rationale generation and optimization. 
Here, we denote the $f_G$ as an actor and the $f_P$ as a critic. 

\noindent \textbf{State-action value and state value learning}. The actor $f_G$ and the critic $f_P$ collaboratively optimize the rationale generation policy, where $f_G$ is responsible for generating rationale candidates (generation policy), and $f_P$ is responsible for estimating the policy and guiding the optimization of the policy.
Formally, the state-action value function $Q$ and the state value function $V$ at time step $t$ can be expressed as:
\begin{equation}\label{e4.2.1}
\begin{aligned}
&Q^\pi(s^t,a)=V^\pi(s^{t+1}),\\
&V^\pi(s^t)={R}^t+\mathcal{H}(Y_{s^t},\hat{Y}_{s^t}),
\end{aligned}
\end{equation}
where $Q(s,a)$ measures the expected return when taking action $a$ in state $s$ under policy $\pi$; $V(s)$ measures the expected return when following policy $\pi$ from state $s$. 
In rationalization, the optimization of the actor $Q(s,a)$ and the critic $V(s)$ are mutually dependent. Specifically, $Q(s,a)$ is influenced by $V(s)$ in the next time step, while $V(s)$ is affected by immediate reward ${R}^t$ and supervised label loss $\mathcal{H}$.
Here, ${R}^t$ is the sparsity and continuity constraints (Eq.\ref{Eq3}); 
$\mathcal{H}$ represent the calculated loss by the groundtruth label $Y_{s^t}$ and prediction $\hat{Y}_{s^t}$ at timestep $t$.

\noindent \textbf{Policy update and estimation}. Given a policy $\pi$ at timestep $t$, we can formalize the policy update for actor $f_G$ by computing an advantage function \cite{sutton1999policy} in rationalization.
\begin{equation}\label{e4.2.2}
\begin{aligned} 
\nabla_\theta J(\pi_\theta)&=\mathbb{E}_{s\sim d^\pi,a\sim\pi}\left[\nabla_\theta\log\pi_\theta(a|s)A^\pi(s,a)\right],\\
&=\mathbb{E}_{X\sim d^\pi,Z\sim\pi}\left[\nabla_\theta\log\pi_\theta(Z|X)A^\pi(X,Z)\right],
\end{aligned}
\end{equation}
where $d^\pi(s)$ represents the state distribution under policy $\pi$, and $\nabla_\theta$ denotes the direction of gradient update;
$A^\pi(s^t,a)=Q^\pi(s^t,a)-V^\pi(s^t)$ indicates the advantage of the action $a$ compared to the current return of the state $s$ at timestep $t$.
In rationalization, the state-action function $Q^\pi(s^t,a)$ is estimated by the state-value function $V^\pi(s^{t+1})$ at the next time step, where 
\begin{equation}\label{e4.2.3}
\begin{aligned}
V^\pi_{{s^{t+1} \sim X}}(s^{t+1})=\Omega(M)+\mathcal{H}(Y_{s^{t+1}},\hat Y_{s^{t+1}}),\hat Y_{s^{t+1}}=f_P(s^{t+1}).
\end{aligned}
\end{equation}

\begin{table}[t]
\caption{A toy example payoff (negative entropy) table of the optimization in accordance game, where {Coll. rati.} and {Rati. rati.} represent collapsed rationales and rational rationales, respectively; {Real. loss} indicates $L_{MMI}$ in Eq.\ref{Eq110_}. }
\centering
\begin{tabular}{|l|ccccc|}
\hline
Rati. mode $r$&  Real. loss & $V(s^t)$ & $\epsilon(s^t)$ & $Q(s^t,a)$ & $A(s^t,a)$ \\
\hline
Rati. rati. & 1.0 & 10000.0 & -9999.0 & 9999.4 & -0.6$\approx$0 \\
Coll. rati.(1) & 8500.0 & 1.0 & 8499.0 & 1.3 & 0.3$\approx$0 \\
Coll. rati.(2) & 9000.0 & 1.0 & 8999.0 & 1.1 & 0.1$\approx$0 \\
Coll. rati.(3) & 10000.0 & 1.0 & 9999.0 & 0.9 & -0.1$\approx$0 \\
\hline
\end{tabular}
\label{tab:An example payoff}
\vspace{-4pt}
\end{table}

\noindent \textbf{Coordinated Game-theoretic Problem: The actor who no longer tends to explore new strategies leads to a suboptimal equilibrium.}
To intuitively illustrate the coordinated problem between two players, we first reveal a phenomenon about empirical bias.
As shown in Table \ref{tab:An example payoff}, regardless of whether the given rationale mode $r$ is collapsed, once the critic function $V^\pi(s)$ introduces an erroneous bias in its estimation, it will lead to the advantage function $A^\pi(s,a)$ guiding the actor $Q^\pi(s,a)$ towards a near-zero. Moreover, $Q^\pi(s,a)$ learns this error $\epsilon(s)$ introduced by $V^\pi(s)$.
With Proposition \ref{def:prop}, we have that the actor $f_G$ relies solely on the critic $f_P$'s estimation. 
Errors in the critic's estimation will prevent the policy from converging to the optimal solution.
Formally, we can express error $\epsilon(s)$ at timestep $t$ by
\begin{equation}\label{e4.2.4} 
\begin{aligned}
V^\pi(s)= V^*(s)-\epsilon(s),
\end{aligned}
\end{equation}
where $V^*(s)$ is the optimal state value, and $\epsilon(s)$ represents the estimation error of the critic $f_P$.
Therefore, an incorrect advantage function is generated under a given policy,
\begin{equation}\label{e4.2.5}
\begin{aligned}
A^\pi(s,a)&=Q^\pi(s,a)-V^\pi(s)= Q^*(s,a)-(V^*(s)-\epsilon(s)),\\
&= Q^*(s,a)-V^*(s)+\epsilon(s).
\end{aligned}
\end{equation}
If $\epsilon(s)$ is too large or consistently negative at timestep $t$, then $A^\pi(s,a)$ may become excessively small, preventing the actor $f_G$ from exploring new strategies.
Therefore, we can establish the following theorem.

\begin{theorem}\label{thm:ns}
Given an RNP model with $f_G$ and $f_P$. 
Let $\epsilon(s^t)$ be the estimation error of the $f_P$ for a candidate rationale $\hat Z_t$ at timestep $t$. 
If exist $\epsilon(s^t)\neq0$, then the $f_G$ no longer tends to explore new strategies to uncover more informative rationale. That is to say, the policy of $f_G$ $\pi$ satisfies that 
\begin{equation}\label{e4.2.6}
\begin{aligned}
&\exists \pi\neq\pi^*, \quad\forall(s^t,a),A^{\pi}(s^t,a)=0 \\
&\Rightarrow \nabla_\theta J(\pi)=\mathbb{E}_{s^t,a}\left[\nabla_\theta\log\pi(a|s^t)A^{\pi}(s^t,a)\right]=0.
\end{aligned}
\end{equation}
\end{theorem} 
\noindent Theorem \ref{thm:ns} suggests that \textit{regardless of the policy profile of $f_G$, if the estimation of $f_P$ is biased, the RNP model will no longer tends to explore new strategies, which leads to a continual degeneration}.

\section{Methodology and Theoretical Analysis}\label{method}

To address the above problem, we propose PORAT (Fig.\ref{fig:PORAT}), a game-theoretic policy optimization for self-explanation rationalization, including {the proposed method} (Sect. \ref{5.1}) and theoretical insights (Sect. \ref{5.2}). 

\subsection{The Proposed Method}\label{5.1}
As shown in Equation \ref{e4.2.4}, intuitively, if we introduce a regularization penalty term, 
the error of the critic can be alleviated. 
Recent work \cite{Huang2021DMR,yu2021A2R,liu2022fr,yue2022dare,liu2024enhancing} has explored this through calibrating or penalizing the predictor. 
However, the penalty factor is difficult to control, which could lead to longer optimization paths or introduce extra local optima \cite{liu2024mrd}.
When the model converges to local optima, these approaches also encounter a bottleneck.

Diverging from previous research, we aim to develop a method to help the model cope with such continual degeneration so that, 
regardless of the strategy chosen by the model, gradient-based descent can guide it out of local optima. 
We assume that at timestep $t$, the RNP model is in a suboptimal state. According to Theorem \ref{thm:ns}, we can derive that the $f_G$'s policy gradient is nearly zero,
\begin{equation}\label{e5.1}
\nabla_{\theta}J(\pi)=\mathbb{E}_{s^t,a}\left[\nabla_{\theta}\log\pi(a|s^t)A^{\pi}(s^t,a)\right]=0,
\end{equation}
which means that the $f_G$ no longer explores new actions, falling into a continual degeneration.  
Furthermore, we have 
\begin{equation}\label{e5.2}
A^{\pi}(s^t,a)=Q^{\pi}(s^t,a)-V^*(s^t)+\epsilon(s^t)=0.
\end{equation}
If we can ensure that $A^{\pi}(s^{t+1},a)\neq0$, $f_G$ will be able to escape the suboptimal equilibrium at time $t+m$ ($m>0$). Formally, this can be expressed as:
\begin{equation}\label{e5.3}
A^{\pi}(s^{t+1},a)=Q^{\pi}(s^{t+1},a)-V^{\pi}(s^{t+1})\neq 0. 
\end{equation}
We first need to confirm whether there is a more optimal policy selection.
Here, we establish the following lemma. 
\begin{lemma}\label{thm:5.1}
Let $S=\{X_1,\dots,X_{2^l}\}$ be the set of all candidate rationales for a given input $X$, and let $+C$ and $-C$  suggest a best rationale and a suboptimal one. Suppose $f_G$ satisfies a suboptimal state $s^t$ at timestep $t$, there exists at least one state induced by the corresponding policy profile $\pi(a|s^{t+1})$ that enables $f_G$ to escape the $s^t$, that is,
\begin{equation}\label{e5.4}
\begin{aligned}
\forall s^t\sim d^\pi \in S, \nabla_{\theta}J(\pi)=0 
\Rightarrow \exists \pi(a|s^{t+1}), s.t.\nabla_{\theta}J(\pi)\neq 0,
\end{aligned}
\end{equation}
and $\pi(a|s^{t+1})=\{\pi^G_{+C}\times\pi^P_{-C}\}$ and $\{\pi^G_{-C}\times\pi^P_{+C}\}$ are two solutions for the policies of $f_G$ and $f_P$.
\end{lemma} 
\noindent Lemma \ref{thm:5.1} means that:
there exists a strategy $\pi_{t+1}$ that enables the model to escape the suboptimal state $s^t$, 
and the policy $\pi_{t+1}$ is from $\pi^i_{j}$($i \in \{f_G,f_P\},j \in \{+C,-C\}$).
However, according to the Proposition \ref{def:prop}, we have the non-discriminability for predictor, which means if $f_G\Rightarrow R$, then $R \Rightarrow f_P$.

\noindent \textbf{Parameter Freezing as Intervention}. To this end, we disentangle the game between the generator and the predictor from the policy optimization perspective, as shown in Fig.\ref{fig:PORAT}.
Specifically, we first freeze the generator while keeping the predictor active, which allows the generator to block the predictor's 
suboptimal feedback and generate diverse candidate rationales as optional strategies. Formally, let $V^{\pi}(s^{t+1})=0$, we can rewrite the Equation \ref{e5.3} as
\begin{equation}\label{e5.5}
A^{\pi}(s^{t+1},a)=Q^{\pi}(s^{t+1},a)-V^{\pi}(s^{t+1})=Q^{\pi}(s^{t+1},a)\neq0
\end{equation}
Since the model is in a suboptimal state at timestep $t$, Equation \ref{e5.5} is equivalent to the generator selecting suboptimal rationale at time $t+1$, 
while the predictor does not further fit it. In addition, we have $A^{\pi}(s^{t+1},a)\neq0$, allowing the generator to continue exploring new actions.
However, by continuously optimizing Equation \ref{e5.5}, the error induced by the predictor's estimation will be learned by the new $Q^{\pi}(s^{t+1},a)$. Therefore, we further freeze the predictor to mitigate the impact of errors arising from the suboptimal state. This allows the predictor to block the continuously degenerating parameter updates.
According to Equation \ref{e4.2.1}, we have $Q^\pi(s^t,a)=V^\pi(s^{t+1})$, so,
\begin{equation}\label{e5.6} 
A^{\pi}(s^{t+1},a)=Q^\pi(s^t,a)-V^\pi(s^t)=V^\pi(s^{t+1})-V^{\pi}(s^t).
\end{equation}
Intuitively, if $V^\pi(s^{t+1})-V^{\pi}(s^t)=0$, then $f_P$ will overfit the state $s^t$.
Therefore, to address the problem, we let $V^\pi(s^{t+1})=0$, and freeze the predictor $f_P$ at timestep $t+1$ in practice.
Finally, following the general setup of RNP, we simultaneously activate both the $f_G$ and the $f_P$, enabling them to collaborate once again.

\noindent \textbf{Policy Optimization}.
Based on above, given the input $X$ at timestep $t+1$, the learning objective of the model $J(\pi)$ can be represented as
\begin{equation}\label{e5.7}
\begin{aligned} 
&J(\pi(a|s^{t+1}))=
\mathbb{E}^{(1)}_{s\sim d^\pi,a\sim\pi}\left\{\log\pi_\theta(a|s^{t+1})[Q^{\pi}(s^{t+1},a)]\right\}\\
&+ \mathbb{E}^{(2)}_{s\sim d^\pi,a\sim\pi}\left\{\log\pi_\theta(a|s^{t+1})[-V^{\pi}(s^{t})]\right\}\\
&+ \mathbb{E}^{(3)}_{s\sim d^\pi,a\sim\pi}\left\{\log\pi_\theta(a|s^{t+1})[Q^{\pi}(s^{t+1},a)-V^{\pi}(s^{t+1})]\right\}\\
\end{aligned} 
\end{equation} 
further derivation, we have 
\begin{equation}\label{e5.8}
\begin{aligned} 
J(\pi_{\theta})
&=\min_{\theta_p^{t+1}} \mathcal{H}(Y,f^t_P(f^t_G(X^{t+1}))+\Omega(M)\\
&+\min_{\theta_g^{t+1}} \mathcal{H}(Y,f^{t+1}_P(f^t_G(X^{t+1})))+\Omega(M)\\
&+\min_{\theta_g^{t+1},\theta_p^{t+1}} \mathcal{H}(Y,f^{t+1}_P(f^{t+1}_G(X^{t+1})))+\Omega(M)
\end{aligned}
\end{equation}
\noindent \textbf{Iteration and Inference}. 
Equation \ref{e5.8} ensures that the model can temporarily adjust a state. However, the local minima explored by multilayer networks are not unique \cite{kawaguchi2016deep,choromanska2015loss}.
To address this issue, we introduce a progressive optimization process \cite{karras2018progressive} and set the update timestep to $N$.
After $N$ time steps, we reintroduce the aforementioned policy optimization for model iteration. The final optimization process can be expressed as follows:
\begin{equation}\label{e5.9}
L=\sum_{i=0,i\notin\{k*N\}}^T L_1(Y,\hat Y)+\sum_{i\in\{k*N\}}^T L_2(Y,\hat Y), k=\{1,\dots,n\}
\end{equation}
where $L_1(Y,\hat Y)=\min_{\theta_g,\theta_p} \mathcal{H}(Y,\hat{Y}) + \Omega(M)$, $L_2(Y,\hat Y)=J(\pi^{k*N}_\theta)$.
During the inference phase, following the general self-explanation setup \cite{lei2016rnp,liu2023decoupled,zhao-etal-2024-agr}, the $f_P$ only uses the rationale generated by the $f_G$ for model prediction.

\noindent \textbf{Algorithm.} To help readers better understand the process, we detail the main steps for the training and inference phase in Algorithm \ref{alg:PORAT}, in which the input is a dataset $\mathcal{D}$, and the output is a model $\theta_f^*$ capable of providing both predictions $\hat{y}_j$ and rationales $\hat{z}_j$ for a data sample $x_j$.

\begin{algorithm}[h]
   \caption{PORAT Algorithm}
   \label{alg:PORAT}
\begin{algorithmic}[1]
   \STATE {\bfseries Input:} A dataset $\mathcal{D}$, including $\mathcal{D}_{\mathrm{train}}$ and $\mathcal{D}_\mathrm{test}$ \label{ln:1}
   \STATE {\bfseries Output:} A self-explanation model $\theta_f^*$ \label{ln:2}
	\STATE // Training Phase
   \WHILE{not converged} 
       \FOR{$x_{i}\in \mathcal{D}_{\mathrm{train}}$, \textbf{in epoch}} 
           \STATE Compute the output of $f_\mathrm{G}$ and $f_\mathrm{P}$:  
			\begin{equation}\label{Eqa_1}
             \hat y_{i} = f_\mathrm{P}(\theta_\mathrm{p}^{i},f_\mathrm{G}(\theta_\mathrm{g}^{i},x_{i})).
			\vspace{-6pt}
			\end{equation} 
			\STATE Minimize $\mathcal{H}(y_i,\hat{y}_i)$.
           \STATE Update $\theta_\mathrm{p}^{i},\theta_\mathrm{g}^{i}$ through gradient decent.\label{ln:8}
		   \STATE $\theta_\mathrm{g}^{*} \leftarrow \theta_\mathrm{g}^{i}$; $\theta_\mathrm{p}^{*}\leftarrow \theta_\mathrm{p}^{i}$
		   \STATE // Game-theoretic Policy Intervention
           \STATE Execute Eq.(\ref{e5.5}): freezing $\theta_\mathrm{g}^{i}$, update $\theta_\mathrm{p}^{i}$.
		   \STATE Execute Eq.(\ref{e5.6}): freezing $\theta_\mathrm{p}^{i}$, update $\theta_\mathrm{g}^{i}$.
		   \STATE Update $\theta_\mathrm{g}^{i}$, update $\theta_\mathrm{p}^{i}$.

       \ENDFOR  
	\ENDWHILE 
	\STATE // Inference Phase 
	\FOR{$x_{j}\in \mathcal{D}_{\mathrm{test}}$} 
           \STATE Compute the predicting label $\hat y_j$ and generated explanation $\hat z_j$ using Eq.(\ref{Eqa_1}) by substituting the parameters $\theta_\mathrm{g}^*$ and $\theta_\mathrm{p}^*$.
   \ENDFOR 
	\STATE {\bfseries Output:} $\theta_f^* = \theta_\mathrm{g}^* \bigcup \theta_\mathrm{p}^*$; $(x_j,\hat{z}_j,\hat{y}_j),\forall x_j\in D_\mathrm{test}$
\end{algorithmic} 
\end{algorithm}

\vspace{-2pt}
\subsection{Theoretical Analysis}\label{5.2}
\vspace{-2pt}

\begin{theorem}\label{thm:ns2}
Given an RNP model with two players (i.e., $f_G$ and $f_P$) and a suboptimal state $s^t$ at timestep $t$, where 
$s^t$ indicates an collapsed rationale candidate $\hat Z_t \in X $ from $f_G$ at time $t$.
Suppose $f_G$ satisfies the following condition,
\begin{equation}\label{Eq15}
\nabla_\theta J(\pi)=\mathbb{E}_{s,a}\left[\nabla_\theta\log\pi(a|s^t)A^{\pi}(s^t,a)\right]=0,
\end{equation}
then, we have that after $m$ time steps, the fusion and optimization of additional policy actions can be such that
\begin{equation}\label{Eq15}
A^{\pi}(s^{t+m},a)=Q^{\pi}(s^{t+m},a)-V^{\pi}(s^{t+m})\neq 0.
\end{equation}
\end{theorem} 

\noindent \textbf{\textit{Proof.}} We first denote a suboptimal state as $s^t$. According to Theorem \ref{thm:ns}, we can obtain 
\begin{equation}\label{11}
\nabla_\theta J(\pi)=\mathbb{E}_{s,a}\left[\nabla_\theta\log\pi(a|s^t)A^{\pi}(s^t,a)\right]=0,
\end{equation}
which means $A^{\pi}(s^t,a)=0$, the generator $f_G$ no longer tends to explore new strategies.
According to Theorem 1, we define the gain change function of both players as $\varphi_{i,a_i}(\pi)=\max\{0,J_i(a_i,\pi_{-i})-J_i(\pi)\}$ where $i$ indicates the $i$-th player.
When $\varphi_{i,a_i}(\pi)>0$, 
\begin{equation}\label{Eq15}
\begin{aligned}
&\max\{0,J_i(a_i,\pi_{-i})-J_i(\pi)\}>0,\\
&J_i(a_i,\pi_{-i})>J_i(\pi),
\end{aligned}
\end{equation}
we have $A^{\pi}(s^t,a)\neq0$. 
However, according to Theorem \ref{thm:ns}, the joint action candidate of $f_G$ and $f_P$ under the RNP game cannot find a better joint policy that improves the payoff of the RNP model.
Therefore, we need to introduce additional policies to enable the model to find the global optimum $s^*$, but finding the global optimum directly is difficult \cite{yun2018global}.
According to Lemma \ref{thm:5.1}, we can identify at least two intermediate policy points that guide the model to escape the suboptimal state.

Therefore, if we can integrate these two policy profiles, we will be able to guide the advantage function such that 
$A^{\pi}(s^t,a)\neq0$. 
Furthermore, with Equation \ref{e4.2.1}, we have
\begin{equation}\label{e16}
A^{\pi}(s^{t+1},a)=V^\pi(s^{t+1})-V^{\pi}(s^t).
\end{equation}
This indicates that we can learn the aforementioned intermediate policy points in the RNP game by controlling the value function $V^{\pi}(s)$, such that,
\begin{equation}\label{Eq15}
A^{\pi}(s^{t+m},a)=Q^{\pi}(s^{t+m},a)-V^{\pi}(s^{t+m})\neq 0.
\end{equation}
where $m$ indicates the learning period. The proof of Theorem \ref{thm:ns2} is finished, which suggests that we can let the RNP models learn policy optimization to solve the suboptimal state for self-explanation rationalization.

\vspace{-2pt}
\section{Experiments}\label{experiments}
\vspace{-2pt}
In this section, we evaluate our method PORAT in various settings to demonstrate its effectiveness.
\vspace{-2pt}
\subsection{Experimental Setup}
\vspace{-2pt}
\noindent \textbf{Datasets.}
We compare PORAT using BeerAdvocate \cite{beer}, BeerAdvocate*\cite{lei2016rnp} and HotelReview \cite{hotel}, which are three multi-aspect sentiment classification benchmarks widely used in rationalization. 
Note that each of them contains three distinct aspects, which are trained independently in our experiments. Consequently, these three benchmarks can be considered as nine distinct datasets to some extent.
Following previous research \cite{Huang2021DMR,zhao-etal-2024-agr,liu2023decoupled}, we obtain BeerAdvocate \cite{beer}, BeerAdvocate*\cite{lei2016rnp}, and HotelReview \cite{hotel} datasets, which are all publicly available.
As shown in Table \ref{tab:dataset}, the specific splitting details of the nine datasets are presented.
In particular,  BeerAdvocate is a correlated dataset on beer reviews that can be regarded as addressing the spurious correlation problem, 
while BeerAdvocate* is a dataset decorrelated by Lei et al. \cite{lei2016rnp} that focuses on the degeneration problem. 
For HotelReview, it is another benchmark also widely used in rationalization. In synthetic settings, we use the same experiment setup as Yu
et al. \cite{yu2021A2R}, Liu et al. \cite{liu2023decoupled} and Wu et al. \cite{wudiscovering} did.

\begin{table}[t]
  \centering
  \caption{Statistics of datasets where Pos and Neg denote the number of positive and negative examples in each set.}
  \resizebox{0.99\columnwidth}{!}{  
    \begin{tabular}{|c l| c c| c c| c c |c|}
    \hline
         \multirow{2}{*}{Benchmarks}&\multirow{2}{*}{Datasets}&\multicolumn{2}{c|}{Train}&\multicolumn{2}{c|}{Dev}&\multicolumn{2}{c|}{Annotation}  \\
         \multicolumn{2}{|c|}{}& Pos&Neg&Pos&Neg&Pos&Neg\\
         \hline
        \multirow{3}{*}{BeerAdvocate \cite{beer}}&Appearance&202385&12897 &28488&1318&923&13\\
        {}&Aroma&172299&30564&24494&3396&848&29\\
        {}&Palate&176038&27639&24837&3203&785&20\\
        \hline
        \multirow{3}{*}{BeerAdvocate* \cite{lei2016rnp}}&Appearance*&16891&16891 &6628&2103&923&13\\
        {}&Aroma*&15169&15169&6579&2218&848&29\\
        {}&Palate*&13652&13652&6740&2000&785&20\\
        \hline
        \multirow{3}{*}{HotelReview \cite{hotel}}&Location&7236&7236 &906&906&104&96\\
        {}&Service&50742&50742&6344&6344&101&99\\
        {}&Cleanliness&75049&75049&9382&9382&99&101\\
        \hline
    \end{tabular}
    }
    \vspace{-10pt}
    \label{tab:dataset}
\end{table}

\noindent \textbf{Baselines.}
To validate the effectiveness of PORAT in a rationalization framework, 
we compare with seven latest methods for BeerAdvocate, 
including one standard rationalization method: {RNP} \cite{lei2016rnp};
two {calibration-based methods:}  {DARE} \cite{yue2022dare}, {FR} \cite{liu2022fr};
two {causal-based methods:}  {INVRAT} \cite{chang2020invariant}, {MCD} \cite{liu2023mcd};
and two recent {guidance-based methods:} {AGR} \cite{zhao-etal-2024-agr} and {G-RAT} \cite{g-rat}. 
For BeerAdvocate* and HotelReview benchmarks, we compare with one standard method and five recent models, including {RNP} \cite{lei2016rnp}, {DMR} \cite{Huang2021DMR}, {A2R} \cite{yu2021A2R}, {FR} \cite{liu2022fr}, {DR} \cite{liu2023decoupled} and {G-RAT} \cite{g-rat}. 

\noindent \textbf{Evaluation Metrics.} 
Following previous methods \cite{Huang2021DMR,yu2021A2R,yue-etal-2023-interventional,liu-etal-2023-mgr}, we focus on the quality of rationales and adopt Precision (P), Recall (R), and F1-score (F1) as metrics.  To fairly compare, we perform the best results on the validation set before testing on the test set. Here, \texttt{Acc} denotes the accuracy of the prediction task based on the selected rationales, while $S$ represents the average ratio of selected tokens to the total length of the original text.

\noindent \textbf{Implementations.} 
We utilize one-layer 200-dimension bi-directional gated recurrent units (GRUs) \cite{cho2014gru} followed by one linear layer for each of the players, and the word embedding is 100-dimension Glove \cite{pennington2014glove}. We use Adam \cite{adam} as the optimizer. The reparameterization trick for binarized sampling is Gumbel-softmax, which is also the same as existing research \cite{liu2023decoupled,liu2023mcd,bao-etal-2018-deriving}.
To verify the effectiveness of our PORAT by intervening policy, we perform ablation studies by intervening in the policies of both the prefix generator and the prefix predictor. To minimize the influence of other factors, we conduct the ablation experiments using the same hyperparameters as the baseline.
In experiments, we use two different architectures (DR \cite{liu2023decoupled} and AGR \cite{zhao-etal-2024-agr}) as the backbone models to validate PORAT on different benchmarks, respectively.
All our experiments are run on NVIDIA RTX 6000 Ada GPUs with 48GB.

\begin{table*}[t]
\caption{Comparison with previous methods on BeerAdvocate \cite{beer} benchmark. $S$ is a hyperparameter, which encourages that the percentage of the tokens being generated as rationales is close to a pre-defined level. The \Fst{bold} numbers are the best results for our proposed method. The same applies below. } 
\centering

\resizebox{1.99\columnwidth}{!}{
\begin{tabular}{|c c c |c c| c c |c|c c| c c |c |c c| c c |c |}
\hline
\multicolumn{3}{|c|}{\multirow{2}{*}{Methods}} & \multicolumn{5}{c|}{BeerAdvocate-Appearance} & \multicolumn{5}{c|}{BeerAdvocate-Aroma} & \multicolumn{5}{c|}{BeerAdvocate-Plate}\\
\cline{4-18}
\multicolumn{3}{|c|}{} &S& Acc & P & R &\multicolumn{1}{c|}{F1} &S& Acc & P & R &\multicolumn{1}{c|}{F1} &S& Acc& P & R &\multicolumn{1}{c|}{F1}\\
\hline

\multicolumn{3}{|c|}{RNP}  & 10.0 & \multicolumn{1}{c|}{-} & 32.4 & 18.6& 23.6   & 10.0   & \multicolumn{1}{c|}{-}  & 44.8  & 32.4 & 37.6& 10.0 & \multicolumn{1}{c|}{-}& 24.6    & 23.5 & 24.0  \\
\multicolumn{3}{|c|}{INVRAT}&10.0&\multicolumn{1}{c|}{-} &42.6&31.5&36.2& 10.0 & \multicolumn{1}{c|}{-}&41.2&39.1&40.1& 10.0 & \multicolumn{1}{c|}{-}&34.9&45.6&39.5 \\
\multicolumn{3}{|c|}{DARE}&10.0&\multicolumn{1}{c|}{-} &63.9&42.8&51.3&10.0& \multicolumn{1}{c|}{-}&50.5&44.8&47.5&10.0& \multicolumn{1}{c|}{-}&33.1&45.8&38.4 \\
\multicolumn{3}{|c|}{FR} & 11.1 & \multicolumn{1}{c|}{75.8}& 70.4& 42.0 & 52.6& 9.7  & \multicolumn{1}{c|}{87.7}& 68.1 & 42.2 & 52.1& 11.7 & \multicolumn{1}{c|}{87.9} & 43.7& 40.9& 42.3 \\
\multicolumn{3}{|c|}{MCD} & 9.5  & \multicolumn{1}{c|}{81.5}& \underline{94.2} & 48.4& 63.9& 9.9  & \multicolumn{1}{c|}{87.5} & \underline{84.6} & 53.9& 65.8& 9.4  & \multicolumn{1}{c|}{87.3}& 60.9 & 47.1& 53.1\\
\multicolumn{3}{|c|}{AGR} & 12.4 & \multicolumn{1}{c|}{81.3} &80.4&55.4&65.6& 12.3 & \multicolumn{1}{c|}{87.8}&68.4&54.1&60.4& 12.4& \multicolumn{1}{c|}{86.2}&54.4&55.9&55.1 \\ 
\multicolumn{3}{|c|}{G-RAT}&10.5 & \multicolumn{1}{c|}{82.4}  &81.8 &46.3&59.1 &10.5& \multicolumn{1}{c|}{85.2} &82.0&55.4&66.2&9.5& \multicolumn{1}{c|}{89.2} &56.2 & 43.1& 48.8 \\
\hline
\multicolumn{3}{|c|}{PORAT (Ours)} & 13.9 & \multicolumn{1}{c|}{{83.9}} & {79.1} & \NFFst{59.6} & \Fst{68.0} & 11.5 & \multicolumn{1}{c|}{{87.5}} & {80.2} & \NFFst{59.0} & \Fst{68.0} & 11.6 & \multicolumn{1}{c|}{{88.0}} & \NFFst{65.4} & \NFFst{61.1} & \Fst{63.2} \\ \hline\hline  

\multicolumn{3}{|c|}{RNP} & 20.0   & \multicolumn{1}{c|}{-} & 39.4   & 44.9    & 42.0     & 20.0 & \multicolumn{1}{c|}{-}  & 37.5   & 51.9  & 43.5  & 20.0   & \multicolumn{1}{c|}{-} & 21.6 & 38.9    & 27.8  \\
\multicolumn{3}{|c|}{INVRAT}&20.0&\multicolumn{1}{c|}{-} &58.9&67.2&62.8& 20.0 & \multicolumn{1}{c|}{-}&29.3&52.1&37.5& 20.0 & \multicolumn{1}{c|}{-}&24.0&55.2&33.5 \\
\multicolumn{3}{|c|}{DARE}&20.0&\multicolumn{1}{c|}{-} &63.7&71.8&67.5&20.0& \multicolumn{1}{c|}{-}&41.0&61.5&49.3&20.0& \multicolumn{1}{c|}{-}&24.4&54.9&33.8 \\
\multicolumn{3}{|c|}{FR}  & 20.9 & \multicolumn{1}{c|}{84.6}  & 74.9   & 84.9   & 79.6  & 19.5 & \multicolumn{1}{c|}{89.3} & 58.7   & 73.3   & 65.2    & 20.2 & \multicolumn{1}{c|}{88.2}  & 36.6   & 59.4  & 45.3  \\
\multicolumn{3}{|c|}{MCD}  & 20.0   & \multicolumn{1}{c|}{85.5}  & 79.3   & 85.5  & 82.3  & 19.3 & \multicolumn{1}{c|}{88.4}   & 65.8   & 81.4    & 72.8   & 19.6 & \multicolumn{1}{c|}{87.7}  & 41.3    & 65.0   & 50.5    \\
\multicolumn{3}{|c|}{AGR} &19.7&\multicolumn{1}{c|}{85.2}&83.3&\underline{88.4}&85.8&19.6&\multicolumn{1}{c|}{89.2}&65.7&82.7&73.2&18.0&\multicolumn{1}{c|}{87.0}&45.2&65.6&53.5\\
\multicolumn{3}{|c|}{G-RAT}&19.7& \multicolumn{1}{c|}{85.0} &80.2&85.2&82.6&20.2 & \multicolumn{1}{c|}{88.1}&60.5&78.2&68.2&20.3 & \multicolumn{1}{c|}{86.1} & 38.4 &62.7&47.6 \\
\hline
\multicolumn{3}{|c|}{PORAT (Ours)} & 19.3 & \multicolumn{1}{c|}{{85.4}} & \NFFst{84.6} & {88.1} & \Fst{86.3} &19.2& \multicolumn{1}{c|}{{89.4}} & \NFFst{69.4} & \NFFst{85.3} & \Fst{76.5} & 19.3 & \multicolumn{1}{c|}{{86.7}} & \NFFst{50.6} & \NFFst{78.6} & \Fst{61.6} \\ \hline\hline 

\multicolumn{3}{|c|}{RNP} & 30.0   & \multicolumn{1}{c|}{-} & 24.2   & 41.2  & 30.5  & 30.0  & \multicolumn{1}{c|}{-}    & 27.1    & 55.7   & 36.4   & 30.0   & \multicolumn{1}{c|}{-}    & 15.4   & 42.2    & 22.6  \\
\multicolumn{3}{|c|}{INVRAT}&30.0&\multicolumn{1}{c|}{-} &41.5&74.8&53.4& 30.0 & \multicolumn{1}{c|}{-}&22.8&65.1&33.8& 30.0 & \multicolumn{1}{c|}{-}&20.9&71.6&32.3 \\
\multicolumn{3}{|c|}{DARE}&30.0&\multicolumn{1}{c|}{-} &45.5&80.6&58.1&30.0& \multicolumn{1}{c|}{-}&32.7&68.2&44.2&30.0& \multicolumn{1}{c|}{-}&19.7&70.5&30.8 \\
\multicolumn{3}{|c|}{FR}&29.6 & \multicolumn{1}{c|}{86.4}& 50.6 & 81.4   & 62.3    & 30.8 & \multicolumn{1}{c|}{88.1} & 37.4  & 75.0  & 49.9   & 30.1 & \multicolumn{1}{c|}{87.0}  & 24.5   & 58.8  & 34.6      \\
\multicolumn{3}{|c|}{MCD}  & 29.7 & \multicolumn{1}{c|}{86.7}   & 59.6   & 95.6   & 73.4  & 29.6 & \multicolumn{1}{c|}{90.2}   & 46.1   & 87.5   & 60.4  & 29.4 & \multicolumn{1}{c|}{87.0}  & 30.5   & 72.4  & 42.9    \\
\multicolumn{3}{|c|}{AGR} & 28.0 & \multicolumn{1}{c|}{87.4} &61.6&93.3&74.2& 30.6 & \multicolumn{1}{c|}{89.7}&43.5&85.3&57.6& 30.4 & \multicolumn{1}{c|}{88.3}&32.1&78.5&45.6 \\ 
\multicolumn{3}{|c|}{G-RAT}&29.6  & \multicolumn{1}{c|}{87.2}& 56.0 &89.4 &68.9 &29.8 & \multicolumn{1}{c|}{90.4} &42.4&81.1&55.7& 29.7& \multicolumn{1}{c|}{86.2}  &27.0 &64.4 &38.0\\
\hline
\multicolumn{3}{|c|}{PORAT (Ours)}&28.7&\multicolumn{1}{c|}{{86.1}}& \NFFst{61.9}&\NFFst{95.8}&\Fst{75.2}&28.3& \multicolumn{1}{c|}{{90.3}} & \NFFst{48.9}&\NFFst{88.7}&\Fst{63.0} &29.6&\multicolumn{1}{c|}{{89.3}} & \NFFst{33.1} & \NFFst{79.0} & \Fst{46.7} \\ \hline

\end{tabular}
}

\label{tab:beer}
\end{table*}

\begin{table*}[t]
\caption{Comparison with previous methods on BeerAdvocate* \cite{lei2016rnp} benchmark. }%
\centering

\resizebox{1.99\columnwidth}{!}{
\begin{tabular}{|c c c |c c| c c |c|c c| c c |c |c c| c c |c |}
\hline
\multicolumn{3}{|c|}{\multirow{2}{*}{Methods}} & \multicolumn{5}{c|}{BeerAdvocate*-Appearance*} & \multicolumn{5}{c|}{BeerAdvocate*-Aroma*} & \multicolumn{5}{c|}{BeerAdvocate*-Plate*}\\
\cline{4-18}
\multicolumn{3}{|c|}{} &S& Acc & P & R &\multicolumn{1}{c|}{F1} &S& Acc & P & R &\multicolumn{1}{c|}{F1} &S& Acc& P & R &\multicolumn{1}{c|}{F1}\\
\hline

\multicolumn{3}{|c|}{RNP} &18.2&\multicolumn{1}{c|}{83.3}&73.8&72.7&73.2
&16.0&\multicolumn{1}{c|}{85.2}&64.1&65.9&64.9
&13.0&\multicolumn{1}{c|}{85.2}&60.1&63.1&61.5
\\
\multicolumn{3}{|c|}{DMR} &18.2&\multicolumn{1}{c|}{-}&71.1&70.2&70.7
&15.4&\multicolumn{1}{c|}{-}&59.8&58.9&59.3
&11.9&\multicolumn{1}{c|}{-}&53.2&50.9&52.0
\\
\multicolumn{3}{|c|}{A2R} &18.4&\multicolumn{1}{c|}{83.9}&72.7&72.3&72.5
&15.4&\multicolumn{1}{c|}{86.3}&63.6&62.9&63.2
&12.4&\multicolumn{1}{c|}{81.2}&57.4&57.3&57.4
\\
\multicolumn{3}{|c|}{FR} &18.4&\multicolumn{1}{c|}{87.2}&82.9&82.6&82.8
&15.0&\multicolumn{1}{c|}{88.6}&74.7&72.1&73.4
&12.1&\multicolumn{1}{c|}{89.7}&67.8&66.2&67.0
\\
\multicolumn{3}{|c|}{DR} &18.3&\multicolumn{1}{c|}{81.1}&82.4&81.6&82.0
&15.4&\multicolumn{1}{c|}{86.2}&77.7&76.8&77.2
&12.5&\multicolumn{1}{c|}{85.0}&65.9&66.0&66.0\\
\multicolumn{3}{|c|}{G-RAT} &18.5&\multicolumn{1}{c|}{-}&84.8&\underline{83.2}&84.0 
&15.5&\multicolumn{1}{c|}{-}&\underline{79.1}&74.3&76.6
&12.3&\multicolumn{1}{c|}{-}&63.4&67.2&65.2\\
\hline
\multicolumn{3}{|c|}{PORAT (Ours)} &18.1&\multicolumn{1}{c|}{83.1}&\NFFst{85.2}&{83.1}&\Fst{84.2}  
&15.6&\multicolumn{1}{c|}{88.0}&{77.9}&\NFFst{78.2}&\Fst{78.0}  
&12.5&\multicolumn{1}{c|}{84.0}&\NFFst{69.0}&\NFFst{69.0}&\Fst{69.0}
\\\hline

\end{tabular}
}

\label{tab:beer_}
\end{table*}

\begin{table*}[t]
\caption{Comparison with previous methods on HotelReview \cite{hotel} benchmark. }%
\centering

\resizebox{1.99\columnwidth}{!}{
\begin{tabular}{|c c c |c c| c c |c|c c| c c |c |c c| c c |c |}
\hline
\multicolumn{3}{|c|}{\multirow{2}{*}{Methods}} & \multicolumn{5}{c|}{HotelReview-Location} & \multicolumn{5}{c|}{HotelReview-Service} & \multicolumn{5}{c|}{HotelReview-Cleanliness}\\
\cline{4-18}
\multicolumn{3}{|c|}{} &S& Acc & P & R &\multicolumn{1}{c|}{F1} &S& Acc & P & R &\multicolumn{1}{c|}{F1} &S& Acc& P & R &\multicolumn{1}{c|}{F1}\\
\hline

\multicolumn{3}{|c|}{RNP} & 8.8& \multicolumn{1}{c|}{97.5} & 46.2 &48.2 &47.1 
&11.0 &\multicolumn{1}{c|}{97.5} & 34.2 & 32.9&33.5
&10.5&\multicolumn{1}{c|}{96.0}&29.1&34.6&31.6\\
\multicolumn{3}{|c|}{DMR} & 10.7 & \multicolumn{1}{c|}{-} & 47.5 &60.1 & 53.1 
& 11.6 & \multicolumn{1}{c|}{-} & 43.0 &43.6&43.3
&10.3&\multicolumn{1}{c|}{-}&31.4&36.4&33.7\\
\multicolumn{3}{|c|}{A2R} & 8.5 &\multicolumn{1}{c|}{87.5} &43.1 &43.2 & 43.1 
&11.4 & \multicolumn{1}{c|}{96.5} & 37.3 & 37.2&37.2
&8.9&\multicolumn{1}{c|}{94.5}&33.2&33.3&33.3\\
\multicolumn{3}{|c|}{FR} &9.0&\multicolumn{1}{c|}{93.5}& 55.5&58.9&57.1
&11.5& \multicolumn{1}{c|}{94.5}&44.8&44.7&44.8
&11.0&\multicolumn{1}{c|}{96.0}&34.9&43.4&38.7\\
\multicolumn{3}{|c|}{DR}&10.5&\multicolumn{1}{c|}{93.5}&51.7&{63.7}&57.1  
&11.8&\multicolumn{1}{c|}{96.5}&45.0&50.2&47.5  
&10.3&\multicolumn{1}{c|}{94.5}&38.6&45.1&41.6\\
\multicolumn{3}{|c|}{G-RAT} &10.1&\multicolumn{1}{c|}{-}&56.1&59.3&57.6 
&12.1&\multicolumn{1}{c|}{-}&48.8&44.1&46.3
&11.9&\multicolumn{1}{c|}{-}&41.4&37.3&39.2\\
\hline
\multicolumn{3}{|c|}{PORAT (Ours)}  &10.2&\multicolumn{1}{c|}{94.0}&\NFFst{53.4}&\NFFst{63.1}&\Fst{57.8}
&13.2&\multicolumn{1}{c|}{95.5}&\NFFst{45.0}&\NFFst{51.8}&\Fst{48.2}  
&10.6&\multicolumn{1}{c|}{93.5}&\NFFst{38.7}&\NFFst{46.4}&\Fst{42.2}
\\\hline

\end{tabular}
}

\label{tab:hotel}
\end{table*}

\subsection{Evaluation on Standard Benchmarks} 

\noindent \textbf{(1) Results on BeerAdvocate benchmark \cite{beer}.}
We first set the rationale sparsity $S$ to approximately 10\%, 20\%, and 30\% \cite{yue-etal-2023-interventional,liu-etal-2023-mgr,liu2023mcd}. As shown in Table \ref{tab:beer}, we achieve significant improvements in F1 scores across various aspects, with an increase of up to 8.1\% in the Palate aspect ($s\approx10\%$). 
The significant improvement shows the superiority of our proposed game-theoretic policy optimization for PORAT in solving suboptimal rationalization,
which can help RNP models to explore more optimal policies for games. 

\noindent \textbf{(2) Results on BeerAdvocate* benchmark \cite{lei2016rnp}.} As shown in Table \ref{tab:beer_}, the results on BeerAdvocate* are illustrated, which focuses more on the research problem of decorrelation \cite{liu2023decoupled}. 
We can observe that our proposed method once again obtains the best performance across all three aspects of the decorrelated beer dataset consistently.

\noindent \textbf{(3) Results on HotelReview benchmark \cite{hotel}.} Table \ref{tab:hotel} presents the experimental results on the HotelReview. In this benchmark, we set the rationale sparsity close to the human-annotated rationales. We can find that our proposed method also achieves varying degrees of improvement in the Location, Service and Cleanliness three datasets.

In conclusion, we demonstrate that our proposed method PORAT outperforms the best existing methods in terms of F1 score across nine datasets from three benchmark datasets (fifteen experiment settings), while maintaining competitive accuracy. This highlights that proposed PORAT method not only offers more accurate explanations than the existing methods but also exhibits strong generalizability.

\begin{table*}[t]
\caption{Experimental results that induces degeneration on synthetic settings. “skew$k$” means that the predictor is pre-trained for $k$ epochs. }
    \centering
    \resizebox{2.00\columnwidth}{!}{
	\setlength\tabcolsep{3pt}
    \begin{tabular}{|c |c |c| c c c|c| c c c|c| c c c|c| c c c|c| c c c|}
    \hline

    \multirow{2}{*}{Aspect} &\multirow{2}{*}{Setting}& \multicolumn{4}{c|}{RNP} & \multicolumn{4}{c|}{A2R}& \multicolumn{4}{c|}{FR} & \multicolumn{4}{c|}{DR}& \multicolumn{4}{c|}{PORAT (Ours)} \\
\cline{3-22}
{}&{} &Acc&P & R &F1&Acc & P & R &F1&Acc&P & R &F1&Acc&P & R &F1&Acc&P & R &F1\\
\hline
\multicolumn{1}{|c|}{{\multirow{3}{*}{Aroma*}}} &\multicolumn{1}{c|}{skew10} &82.6&68.5 &63.7 & 61.5 &84.5&{78.3}&70.6&69.2& 87.1 &73.9&{71.7}&{72.8} 
&85.0&\underline{77.3}&75.7&76.5
&86.7&{77.0}&\Fst{80.0}&\Fst{78.5}\\
\multicolumn{1}{|c|}{}&\multicolumn{1}{c|}{skew15} &80.4&54.5& 51.6&49.3 &81.8&58.1&53.3&51.7& 86.7 &{71.3}&{68.0}&{69.6} 
&85.4&76.1&77.2&76.6
&86.6&\Fst{77.1}&\Fst{79.2}&\Fst{78.1}\\
\multicolumn{1}{|c|}{}&\multicolumn{1}{c|}{skew20} &76.8&10.8 & 14.1 &11.0 &80.0&51.7&47.9&46.3&85.5 &{72.3}&{69.0}&{70.6}
&85.5&77.3&76.2&76.8
&86.1&\Fst{77.6}&\Fst{79.6}&\Fst{78.6}\\
\cline{1-22}
\multicolumn{1}{|c|}{{\multirow{3}{*}{Palate*}}} &\multicolumn{1}{c|}{skew10} &77.3&5.6 &7.4 & 5.5 &82.8&50.3&48.0&45.5 &75.8&{54.6}&{61.2}&{57.7}
&85.8&67.7&\underline{68.6}&68.2
&84.7&\Fst{68.3}&{68.2}&\Fst{68.3} \\
\multicolumn{1}{|c|}{}&\multicolumn{1}{c|}{skew15} &77.1&1.2 & 2.5 & 1.3 &80.9&30.2&29.9&27.7&81.7&{51.0}&{58.4}&{54.5}
&83.9&66.3&66.7&66.5
&84.7&\Fst{68.0}&\Fst{68.4}&\Fst{68.2} \\
\multicolumn{1}{|c|}{}&\multicolumn{1}{c|}{skew20} &75.6&0.4 & 1.4 & 0.6 &76.7&0.4&1.6&0.6& 83.1 &{48.0}&{58.9}&{52.9}
&85.0&59.4&62.6&61.0
&85.7&\Fst{66.6}&\Fst{67.6}&\Fst{67.1} \\
\hline
    \end{tabular}
    }
    \label{tab:predskew}
\end{table*}

\begin{table*}[t]
\caption{Experimental results that induces spurious correlation on synthetic settings. Here, following the same setting \cite{wudiscovering}, we report the results of three random seeds across nine distinct setups where "bias=$c$" indicates the distinct biases of spurious correlations to use Spurious-Motif datasets \cite{ying2019gnnexplainer}.}%
\centering
\resizebox{1.99\columnwidth}{!}{
\begin{tabular}{|ccc|c|c|c|c|c|c|c|c|c|c|}
\hline
\multicolumn{3}{|c|}{\multirow{2}{*}{Methods}} & \multicolumn{9}{c|}{Suprious-Motif Datasets} &\multirow{2}{*}{Avg} \\
\cline{4-12}
\multicolumn{3}{|c|}{} & bias=0.1 & bias=0.2 & bias=0.3 &\multicolumn{1}{c|}{bias=0.4}& bias=0.5 & bias=0.6 & bias=0.7 & bias=0.8 & bias=0.9 &  \\
\hline
\multicolumn{3}{|l|}{Attention} &-&-&-&-&18.3\std{±13.0}&-&18.2\std{±1.4}&-&13.4\std{±1.3}&-\\
\multicolumn{3}{|l|}{ASAP} &-&-&-&-&18.8\std{±2.3}&-&18.6\std{±2.7}&-&12.1\std{±2.1}&-\\
\multicolumn{3}{|l|}{Topk Pool} &-&-&-&-&20.7\std{±5.7}&-&21.2\std{±5.6}&-&\std{14.8±1.8}&-\\
\multicolumn{3}{|l|}{SAG Pool} &-&-&-&-&19.8\std{±6.2}&-&20.1\std{±6.4}&-&13.6\std{±1.4}&-\\
\multicolumn{3}{|l|}{DIR} &26.2\std{±1.4}&28.0\std{±2.8}&29.8\std{±3.6}&28.8\std{±3.0}&30.2\std{±3.3}&29.9\std{±2.4}&30.8\std{±1.6}&28.7\std{±4.3}&24.4\std{±1.3}&28.5\std{±2.6}\\
\multicolumn{3}{|l|}{DIR-DR} &\XFst{26.0\std{±3.1} \XFst{$\downarrow$}}&28.2\std{±3.9}&\XFst{29.8\std{±4.3} \XFst{$\downarrow$}}&29.2\std{±4.4}&\XFst{29.1\std{±4.1} \XFst{$\downarrow$}}&\XFst{29.0\std{±1.7} \XFst{$\downarrow$}}&\XFst{28.6\std{±1.3} \XFst{$\downarrow$}}&\XFst{28.1\std{±0.7} \XFst{$\downarrow$}}&\underline{26.3\std{±1.6}}&\XFst{28.2\std{±2.8} \XFst{$\downarrow$}}\\
\hline 
\multicolumn{3}{|l|}{DIR-PORAT} &\Fst{27.2\std{±2.8} \Fst{$\uparrow$}}&\Fst{29.7\std{±1.5} \Fst{$\uparrow$}}&\Fst{31.3\std{±1.2} \Fst{$\uparrow$}}&\Fst{29.9\std{±3.1} \Fst{$\uparrow$}}&\Fst{30.9\std{±1.7} \Fst{$\uparrow$}}&\Fst{30.8\std{±1.8} \Fst{$\uparrow$}}&\Fst{31.3\std{±1.9} \Fst{$\uparrow$}}&\Fst{30.4\std{±2.5} \Fst{$\uparrow$}}&{25.5\std{±1.5}}&\Fst{29.7\std{±1.9} \Fst{$\uparrow$}}\\
\multicolumn{3}{|l|}{w/o p.} &26.2\std{±0.5}&28.1\std{±2.2}&28.9\std{±2.7}&28.3\std{±3.8}&30.3\std{±3.0}&30.6\std{±2.7}&31.1\std{±2.0}&27.7\std{±2.6}&20.3\std{±1.7}&27.9\std{±2.4}\\
\multicolumn{3}{|l|}{w/o g.} &25.2\std{±1.2}&26.5\std{±1.5}&29.4\std{±2.4}&28.4\std{±1.9}&29.9\std{±3.2}&28.1\std{±0.6}&29.0\std{±2.1}&28.6\std{±3.6}&25.2\std{±3.0}&27.8\std{±2.2}\\
\hline
\end{tabular}
}
\label{tab:ls1}
\end{table*}

\begin{figure}[t]
\centering
\subfloat[Appearance]{
    \includegraphics[scale=0.32]{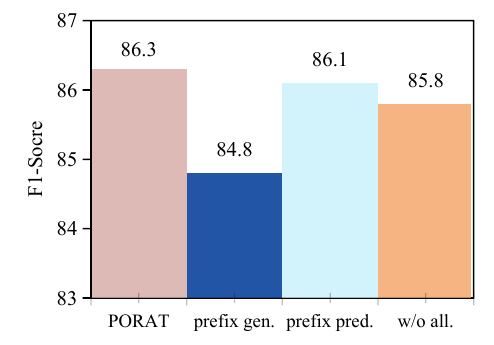} 
}
\subfloat[Aroma]{
    \includegraphics[scale=0.32]{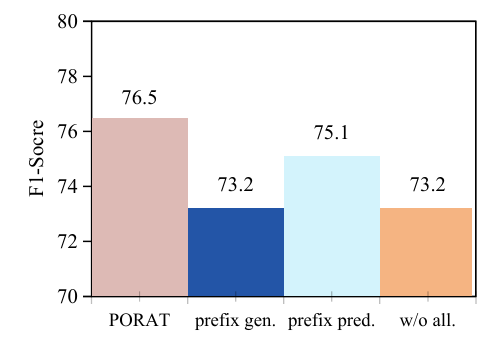}
}
\subfloat[Palate]{
    \includegraphics[scale=0.32]{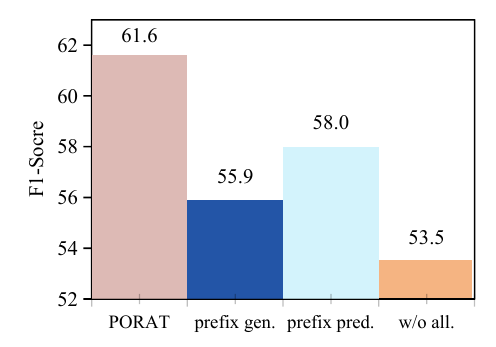}
}
\vspace{-2pt}
\subfloat[Appearance*]{
    \includegraphics[scale=0.32]{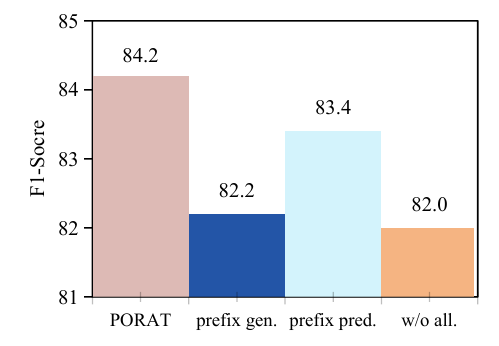}
}
\subfloat[Aroma*]{
    \includegraphics[scale=0.32]{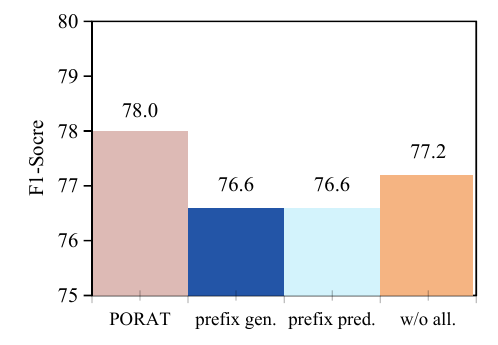}
}
\subfloat[Palate*]{
    \includegraphics[scale=0.32]{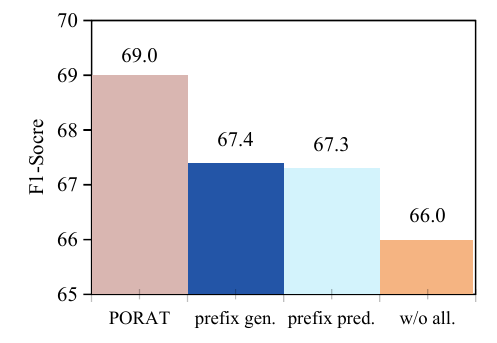}
}
\vspace{-2pt}
\subfloat[Location]{
    \includegraphics[scale=0.32]{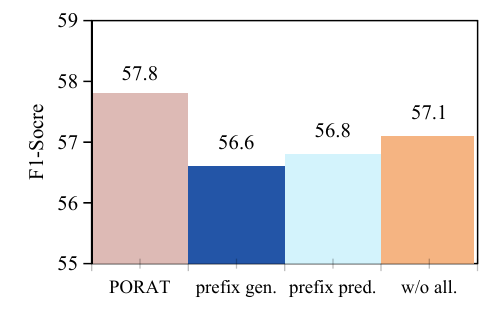}
}
\subfloat[Service]{
    \includegraphics[scale=0.32]{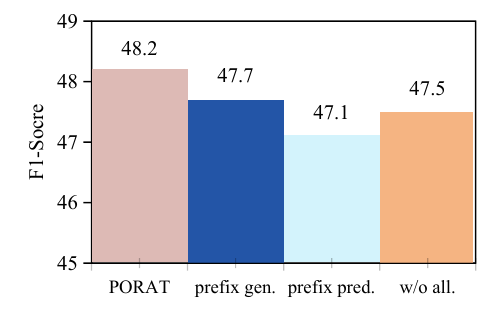}
}
\subfloat[Cleanliness]{
    \includegraphics[scale=0.32]{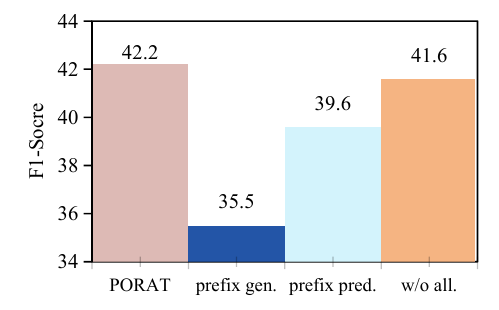}
}

\caption{Ablation Studies on (a-c) BeerAdvocate \cite{beer}, (d-f) BeerAdvocate* \cite{lei2016rnp} and (g-i) HotelReview \cite{hotel} Benchmarks.}
\vspace{-16pt}
\label{figure_ablation}
\end{figure}

\subsection{Evaluation on Synthetic Settings}

To better show the influence of sub-optimal state equilibrium, we further conduct two synthetic experiments.

\noindent \textbf{(1) Degeneration.} It is a typical phenomenon of suboptimal states. To show that even if the predictor overfits to trivial patterns and falls in a suboptimal equilibrium, PORAT can still escape, we conduct the same synthetic experiment as Yu et al. \cite{yu2021A2R} and Liu et al. \cite{liu2023decoupled} did (The specifical setting can be found in Appendices).
The results of inducing degeneration are shown in Table \ref{tab:predskew}. 
We find that PORAT achieves effective improvements across different experimental settings in both aspect-based datasets.
In particular, for the Palate task, under the condition where the skew predictor is trained for 20 epochs, the performance of the previous SOTA model DR, exhibits a degradation trend (68.2 → 66.5 → 61.0).
In contrast, PORAT maintains stable performance (68.3 → 68.2 → 67.1). 
Compared to DR, PORAT achieves up to a 6.1\% improvement in mitigating degeneration.

\noindent \textbf{(2) Spurious correlations.} We also conduct synthetic experiments to evaluate the effectiveness in addressing spurious correlations as Wu et al. \cite{wudiscovering} did (The specifical setting can be found in Appendices). As shown in Table \ref{tab:ls1}, we can observe that, compared to the previous SOTA method DR, PORAT achieves significant improvements in 8 out of 9 experimental settings. In particular, compared to the backbone model, PORAT does not exhibit any performance degradation, whereas DR shows varying degrees of performance decline in six settings (bias=0.1, 0.3, 0.5, 0.6, 0.7, 0.8). This further indicates that DR is sensitive to data distributions with prominent spurious patterns, while PORAT remains robust under such conditions. Moreover, we can see that PORAT also demonstrates more stable performance since it achieves a smaller mean variance.

\begin{table*}[t]
\caption{Results of methods with low sparsity on BeerAdvocate* benchmark. }
\centering

\resizebox{1.99\columnwidth}{!}{
\begin{tabular}{|c c c |c c| c c |c|c c| c c |c |c c| c c |c |}
\hline
\multicolumn{3}{|c|}{\multirow{2}{*}{Methods}} & \multicolumn{5}{c|}{Beer-Appearance*} & \multicolumn{5}{c|}{Beer-Aroma*} & \multicolumn{5}{c|}{Beer-Plate*}\\
\cline{4-18}
\multicolumn{3}{|c|}{} &S& Acc & P & R &\multicolumn{1}{c|}{F1} &S& Acc & P & R &\multicolumn{1}{c|}{F1} &S& Acc& P & R &\multicolumn{1}{c|}{F1}\\
\hline

\multicolumn{3}{|c|}{RNP} & 11.9&-& 72.0 & 46.1 &56.2 &10.7 &- & 70.5 & 48.3&57.3&10.0&-&53.1&42.8&47.5\\
\multicolumn{3}{|c|}{CAR} & 11.9 & - & 76.2 & 49.3 & 59.9 & 10.3 & - & 50.3 &33.3&40.1&10.2&-&56.6&46.2&50.9\\
\multicolumn{3}{|c|}{DMR} & 11.7 &- &{83.6} &52.8 & {64.7} &11.7 &-& 63.1 & 47.6 & 54.3&10.7&-&55.8&48.1&51.7\\
\multicolumn{3}{|c|}{FR} &12.7&83.9 & 77.6&{53.3}&63.2&10.8&87.6& {82.9}&{57.9}&{68.2}&10.0&84.5&{69.3}&{55.8}&{61.8}\\
\multicolumn{3}{|c|}{DR} &12.8&83.8&81.8&56.6&66.9 &11.5&83.5&65.0&60.6&62.7&11.2&82.3&\underline{73.2}&58.6&65.1

\\\hline
\multicolumn{3}{|c|}{PORAT (Ours)} &11.7&84.2&\Fst{90.5}&\Fst{57.3}&\Fst{70.2} &11.6&84.0&\Fst{84.3}&\Fst{62.9}&\Fst{72.0} &10.8&86.7&{71.9}&\Fst{62.4}&\Fst{66.8}
\\\hline

\end{tabular}
}

\label{tab:ls2}
\end{table*}

\begin{table*}[t]
\caption{Experiments on large language model encoder. }
\centering

\resizebox{1.99\columnwidth}{!}{
\begin{tabular}{|c c c |c c c |c|c c c |c |c| c c c |}
\hline
\multicolumn{3}{|c|}{\multirow{2}{*}{Methods}} & \multicolumn{4}{c|}{Hotel-Location} & \multicolumn{4}{c|}{Hotel-Service} & \multicolumn{4}{c|}{Hotel-Cleanliness}\\
\cline{4-15}
\multicolumn{3}{|c|}{} & Acc & P & R &\multicolumn{1}{c|}{F1} & Acc & P & R &\multicolumn{1}{c|}{F1} & Acc& P & R &\multicolumn{1}{c|}{F1}\\
\hline
\multicolumn{15}{|c|}{{In Context Learning (ICL)}}\\
\hline
\multicolumn{3}{|l|}{Llama-3.2-1B (ICL)} &\multicolumn{1}{c|}{74.2}&5.7&5.7&5.7  
&\multicolumn{1}{c|}{83.3}&5.6&5.6&5.6
&\multicolumn{1}{c|}{81.0}&6.2&6.2&6.2\\
\multicolumn{3}{|l|}{Llama-3.2-3B (ICL)} &\multicolumn{1}{c|}{75.9}&7.7&7.9&7.8  
&\multicolumn{1}{c|}{91.3}&10.0&10.0&10.0
&\multicolumn{1}{c|}{91.8}&6.1&6.1&6.1\\
\multicolumn{3}{|l|}{Llama-3.1-8B (ICL)} & \multicolumn{1}{c|}{95.9}&42.8&42.8&42.8  
&\multicolumn{1}{c|}{97.2}&35.9&36.0&36.0
&\multicolumn{1}{c|}{94.3}&23.5&23.6&23.6\\
\hline
\multicolumn{15}{|c|}{{Supervised Fine-Tuning (SFT)}}\\
\hline
\multicolumn{3}{|l|}{Llama-3.2-1B (SFT)} &\multicolumn{1}{c|}{56.8}&11.0&10.8&10.9  
&\multicolumn{1}{c|}{57.6}&11.3&11.5&11.4
&\multicolumn{1}{c|}{57.9}&8.8&8.8&8.8\\
\multicolumn{3}{|l|}{Llama-3.2-3B (SFT)} &\multicolumn{1}{c|}{95.2}&37.4&37.4&37.4  
&\multicolumn{1}{c|}{88.5}&30.2&30.3&30.2
&\multicolumn{1}{c|}{95.5}&17.5&17.5&17.5\\
\multicolumn{3}{|l|}{Llama-3.1-8B (SFT)} &\multicolumn{1}{c|}{84.8}&34.0&34.1&34.0  
&\multicolumn{1}{c|}{90.5}&35.3&35.4&35.4
&\multicolumn{1}{c|}{92.0}&24.9&25.0&25.0\\
\hline
\multicolumn{3}{|l|}{PORAT (Ours)}  &\multicolumn{1}{c|}{94.0}&\Fst{53.4}&\Fst{63.1}&\Fst{57.8}
&\multicolumn{1}{c|}{95.5}&\Fst{45.0}&\Fst{51.8}&\Fst{48.2}  
&\multicolumn{1}{c|}{93.5}&\Fst{38.7}&\Fst{46.4}&\Fst{42.2}
\\\hline

\end{tabular}
}

\label{tab:ls4}
\end{table*}

\begin{table}[t]
\caption{Experiments on small language model encoder. }
\centering

\resizebox{0.99\columnwidth}{!}{
\begin{tabular}{|c c c |c c| c c |c|}
\hline
\multicolumn{3}{|c|}{\multirow{2}{*}{Methods}} & \multicolumn{5}{c|}{Hotel-Location} \\
\cline{4-8}
\multicolumn{3}{|c|}{} &S& Acc & P & R &\multicolumn{1}{c|}{F1} \\
\hline
\multicolumn{3}{|l|}{MCD-BERT-Tiny} & 9.4 & \multicolumn{1}{c|}{85.0 \textcolor{white}{$s$}}&14.7& 16.4&{15.5 \textcolor{white}{$s$}} \\
\multicolumn{3}{|l|}{MCD-BERT-Tiny-OOD} & 10.1& \multicolumn{1}{c|}{89.0 \textcolor{black}{$\uparrow$}} & \XFst{8.5} &\XFst{10.4} &\XFst{9.4 \XFst{$\downarrow$}} \\
\multicolumn{3}{|l|}{MCD-BERT-Tiny-PORAT} &9.8&\multicolumn{1}{c|}{86.5 \textcolor{black}{$\uparrow$}} &\Fst{16.2}&\Fst{18.9}&\Fst{17.4 \Fst{$\uparrow$}}\\

\hline
\multicolumn{3}{|c|}{\multirow{2}{*}{Methods}} & \multicolumn{5}{c|}{Hotel-Service} \\
\cline{4-8}
\multicolumn{3}{|c|}{} &S& Acc & P & R &\multicolumn{1}{c|}{F1} \\
\hline
\multicolumn{3}{|l|}{MCD-BERT-Tiny} &10.7&\multicolumn{1}{c|}{87.5 \textcolor{white}{$s$}}&14.5&13.5&{14.0 \textcolor{white}{$s$}}\\
\multicolumn{3}{|l|}{MCD-BERT-Tiny-OOD} &10.6 &\multicolumn{1}{c|}{94.0 \textcolor{black}{$\uparrow$}} & \XFst{12.1} & \XFst{11.3}&\XFst{11.7 \XFst{$\downarrow$}}\\
\multicolumn{3}{|l|}{MCD-BERT-Tiny-PORAT}&10.6&\multicolumn{1}{c|}{94.5 \textcolor{black}{$\uparrow$}}&\Fst{18.0}&\Fst{18.2}&\Fst{18.1 \Fst{$\uparrow$}}\\

\hline
\multicolumn{3}{|c|}{\multirow{2}{*}{Methods}} & \multicolumn{5}{c|}{Hotel-Cleanliness} \\
\cline{4-8}
\multicolumn{3}{|c|}{} &S& Acc & P & R &\multicolumn{1}{c|}{F1} \\
\hline
\multicolumn{3}{|l|}{MCD-BERT-Tiny} &10.4&\multicolumn{1}{c|}{89.5 \textcolor{white}{$s$}}&17.8&20.6&{19.1 \textcolor{white}{$s$}}\\
\multicolumn{3}{|l|}{MCD-BERT-Tiny-OOD} &10.0&\multicolumn{1}{c|}{97.0 \textcolor{black}{$\uparrow$}} &\XFst{7.8}&\XFst{8.7}&\XFst{8.3 \XFst{$\downarrow$}}\\
\multicolumn{3}{|l|}{MCD-BERT-Tiny-PORAT}&9.6&\multicolumn{1}{c|}{91.0 \textcolor{black}{$\uparrow$}}&\Fst{20.4}&\Fst{21.8}&\Fst{21.1 \Fst{$\uparrow$}}
\\\hline
\vspace{-12pt}
\end{tabular}
}

\label{tab:ls3}
\end{table}

\subsection{Experiment Analysis}

\textbf{(1) Ablation analysis. }
To further verify the effectiveness of PORAT, as well as to investigate the impact of the policies adopted by the predictor and generator players, we remove the optimized policies on three benchmarks to conduct experiments. %
For a fair comparison, we do not specifically tune the hyperparameters and use the same hyperparameters for both PORAT and the modules to be ablated.
As depicted in Fig.\ref{figure_ablation}, by removing all policies, we observe varying degrees of declination across multiply aspects in the nine datasets, 
which highlights the effectiveness of the proposed framework. In addition, additional ablation studies on the spurious correlations synthetic experiments (Table \ref{tab:ls1}) also highlight the importance of both the generator (g.) and the predictor (p.) policy interventions in PORAT.

\begin{figure}[h]
\centering
\subfloat[Appearance*]{
    \includegraphics[scale=0.30]{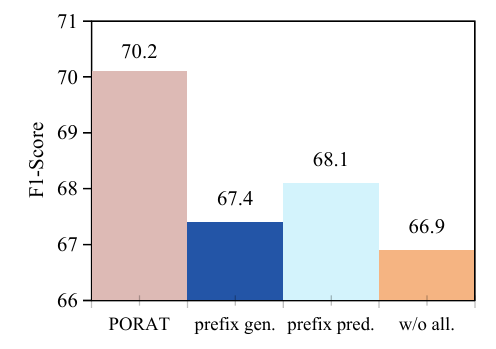}
}
\subfloat[Aroma*]{
    \includegraphics[scale=0.30]{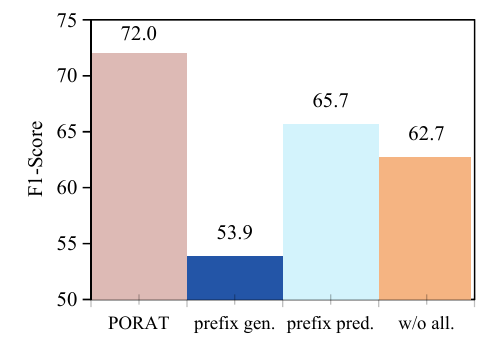}
}
\subfloat[Palate*]{
    \includegraphics[scale=0.30]{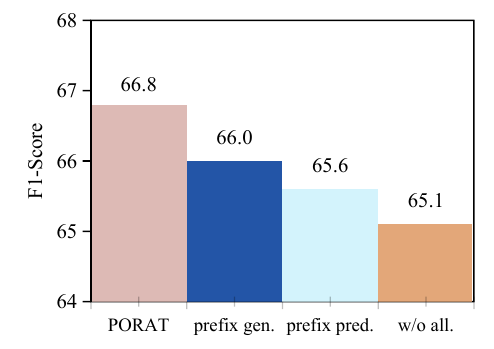}
}
\caption{Low sparsity ablation results.}
\label{figure_low_sparsity}
\vspace{-6pt}
\end{figure}

\noindent \textbf{(2) Low-sparsity analysis. } 
The low-sparsity experiments demonstrate the RNP model's robustness \cite{liu2023decoupled}. Using the same settings \cite{liu2023decoupled,Huang2021DMR,chang2019game}, we also conduct an experiment where the sparsity of selected rationales is extremely low. 
The results are presented in Table \ref{tab:ls2}. We can observe that PORAT also effectively improves the models' robustness compared to previous models.
Besides, we also conduct the ablation results on low-sparsity experiments and observe the impact across multiple aspects (Fig.\ref{figure_low_sparsity}). 
This consistently demonstrates the improvement of PORAT in terms of robustness, which indicates the diversity advantage.

\noindent \textbf{(3) Analysis with pretrained language model encoder. } Then, how do pretrained language models perform in terms of self-explanation? So, we further compare with language models encoders, including fine-tuning (FT)-based, prompt-based and supervised FT (SFT)-based methods, with $\leq$1B, 3B, and 8B parameters. 
As shown in Table \ref{tab:ls4}, we observe that although various types of models can achieve high predictive accuracy, generating self-explainable rationales remains a challenge for language models, and our method outperforms language models with fewer than 8 billion parameters. 
Meanwhile, as shown in Table \ref{tab:ls3}, we also employ language models as backbones to show the competitiveness of our PORAT. 
Consistent with most research \cite{liu2023mcd,liu2025adversarial,liu2025exploring}, here we also use the BERT encoder as a backbone for a fair comparison. 
We observe that our proposed method PORAt not only improves predictive performance but also substantially enhances the self-explanation performance.

\begin{table*}[t]
\caption{Examples of generated rationales from DR and PORAT. Human-annotated rationales are \ul{underlined}. Rationales selected by the generator are highlighted in \textcolor{red}{red}, where "a", "s" and "t" in the input text indicate the rationales annotated by annotators from appearance, aroma and palate aspects.}
\label{tab:case}
\centering
\begin{tabular}{p{8.25cm}p{8.25cm}}
\hline
\textbf{DR} & \textbf{PORAT} \\ \hhline{:=::=:}
\textbf{Aspect:} Beer-Aroma &\textbf{Aspect:} Beer-Aroma  \\
\textbf{Label:} Positive, \textbf{Prediction:} Positive & \textbf{Label:} Positive, \textbf{Prediction:} Positive \\
\textbf{Input Text:} \;\;...\;\; a : pours a clear golden color with a huge 4-finger white head that lasts forever \ul{s :} \textcolor{red}{\ul{spicy phenolic aroma with hints of hops .}} \textcolor{red}{t : smooth heavy malt brew with sweet spices and} alcohol in the background . herbal hops and tad fruity finish . m : medium body and high carbonation . o : very sweet beer - unique - but nothing i would grab again .&
\textbf{Input Text:} \;\;...\;\; a : pours a clear golden color with a huge 4-finger white head that lasts forever \textcolor{red}{\ul{s : spicy phenolic aroma with hints of hops .}} t : smooth heavy malt brew with sweet spices and alcohol in the background . herbal hops and tad fruity finish . m : medium body and high carbonation . o : very sweet beer - unique - but nothing i would grab again .\\
\hline
\hline
\textbf{Aspect:} Beer-Aroma &\textbf{Aspect:} Beer-Aroma  \\
\textbf{Label:} Positive, \textbf{Prediction:} Positive & \textbf{Label:} Positive, \textbf{Prediction:} Positive \\
\textbf{Input Text:} \;\;...\;\; the pour a clear deep amber , the head is mediorce , the lace spare , the color off white . \ul{nose is malt} \textcolor{red}{\ul{, citrus tones , light hints of bubble gum .}} \textcolor{red}{front is malt , sweet ,} the top is medium , the finish is acerbic , dry , the 10 \% abv , is felt in the 'tummy ' and the long lasting alcohol bitter aftertaste . works for me ! , as i like my beers pungent and brawny . ranks \# 504 on my current 1000 beer master list .&
\textbf{Input Text:} \;\;...\;\; the pour a clear deep amber , the head is mediorce , the lace spare , the color off white . \textcolor{red}{\ul{nose is malt , citrus tones , light hints of bubble gum .}} front is malt , sweet , the top is medium , the finish is acerbic , dry , the 10 \% abv , is felt in the 'tummy ' and the long lasting alcohol bitter aftertaste . works for me ! , as i like my beers pungent and brawny . ranks \# 504 on my current 1000 beer master list .\\
\hline
\hline
\textbf{Aspect:} Beer-Aroma &\textbf{Aspect:} Beer-Aroma  \\
\textbf{Label:} Positive, \textbf{Prediction:} Positive & \textbf{Label:} Positive, \textbf{Prediction:} Positive \\
\textbf{Input Text:} \;\;...\;\; reviewed halloween evening , 2009 . poured a very nice deep copper color with fantastic head and lacing . \ul{great scent ,} \textcolor{red}{\ul{very deep bitter aromas , a lot of citrus tones and a slight pine tinge .}} \textcolor{red}{great taste , a nice deep maltiness with} a fantastic bitter ending ; very nice american hops ( citrus ) with a nice earthy undertone to it . goes down very nice , with just the slightest hop roughness . great beer .&
\textbf{Input Text:} \;\;...\;\; reviewed halloween evening , 2009 . poured a very nice deep copper color with fantastic head and lacing . \textcolor{red}{\ul{great scent , very deep bitter aromas , a lot of citrus tones and a slight pine tinge .}} great taste , a nice deep maltiness with a fantastic bitter ending ; very nice american hops ( citrus ) with a nice earthy undertone to it . goes down very nice , with just the slightest hop roughness . great beer .\\
\hline
\end{tabular}
\end{table*}

\subsection{Optimization and Parameter Analysis}
\noindent \textbf{(1) Analysis of different prefix-player policy optimization. }
We also investigate the results of intervening with different prefix player policies. 
We discover that the impact of policy optimization for different players varies across different datasets. 
As shown in Fig.\ref{figure_ablation}, in the BeerAdvocate benchmark, intervening in the prefix predictor of the baseline model effectively helps the model escape the suboptimal state, leading to a significant improvement in model performance (from 85.8/73.2/53.5 to 86.1/75.1/58.0).
However, this is not the case on the other two benchmarks, which may be related to the inherent difficulty of the datasets and their distribution.
The experiments with the out-of-distribution (OOD) method in Table \ref{tab:ls3} further validate that while the distribution improves predictive performance, it also impacts the self-explanatory capability. Therefore, optimizing the strategy for different pattern distributions helps improve performance in rationalization.

\begin{figure}[t]
\vspace{-12pt}
\centering
\subfloat[Aroma*-10]{
    \includegraphics[scale=0.22]{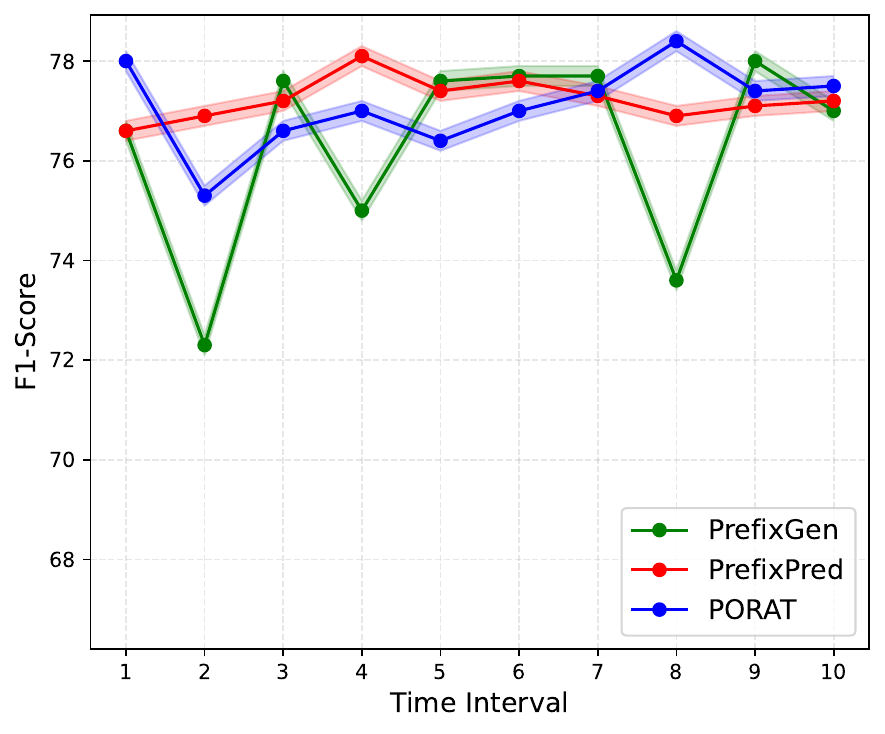}
} 
\subfloat[Palate*-10]{
    \includegraphics[scale=0.22]{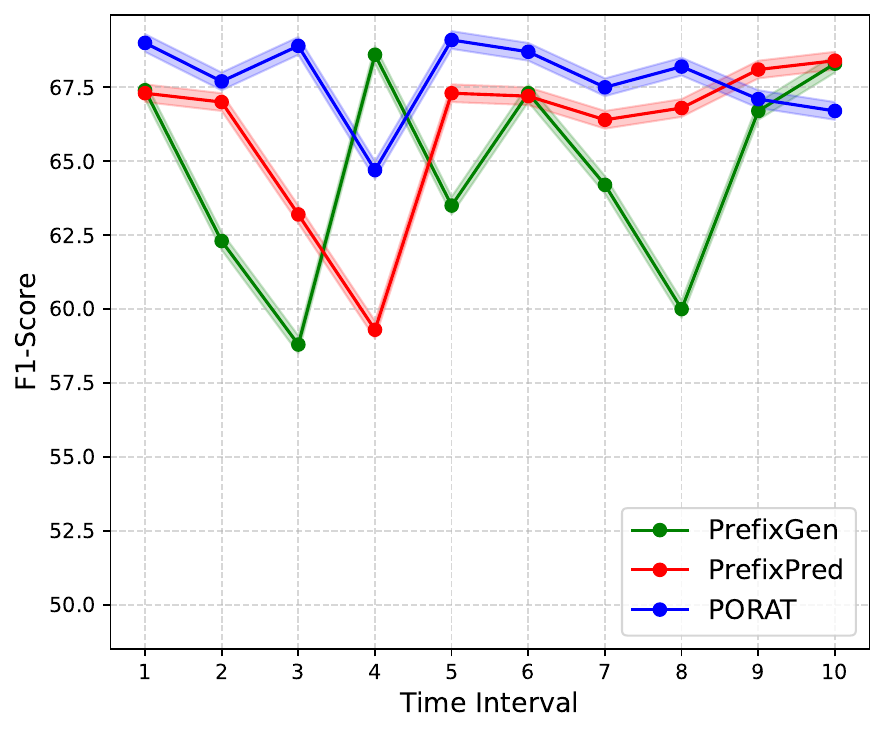}
}
\\
\subfloat[Aroma*-200]{
    \includegraphics[scale=0.22]{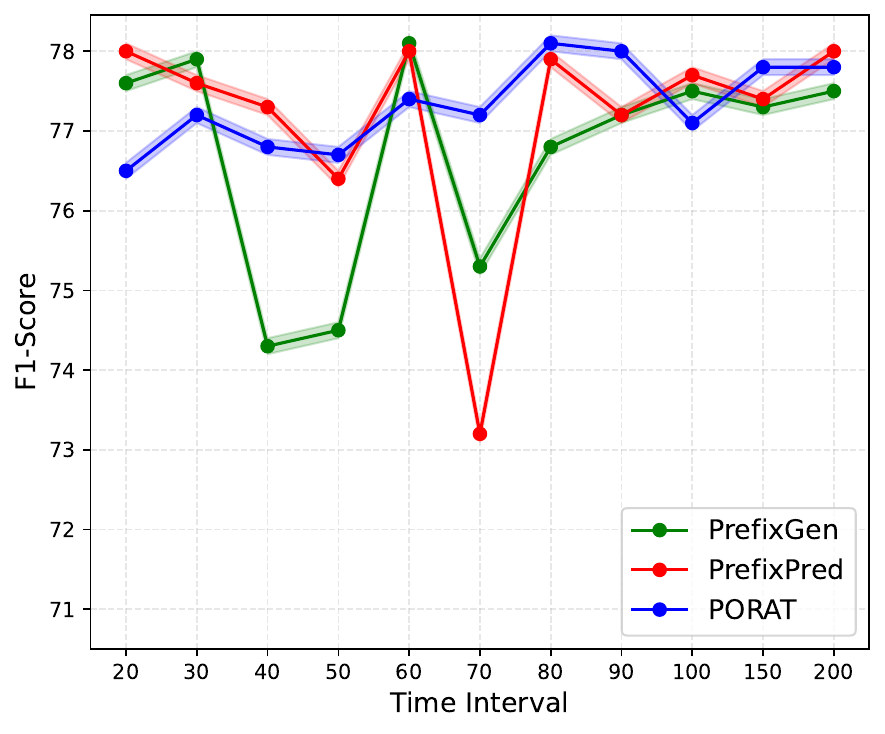}
} 
\subfloat[Palate*-200]{
    \includegraphics[scale=0.22]{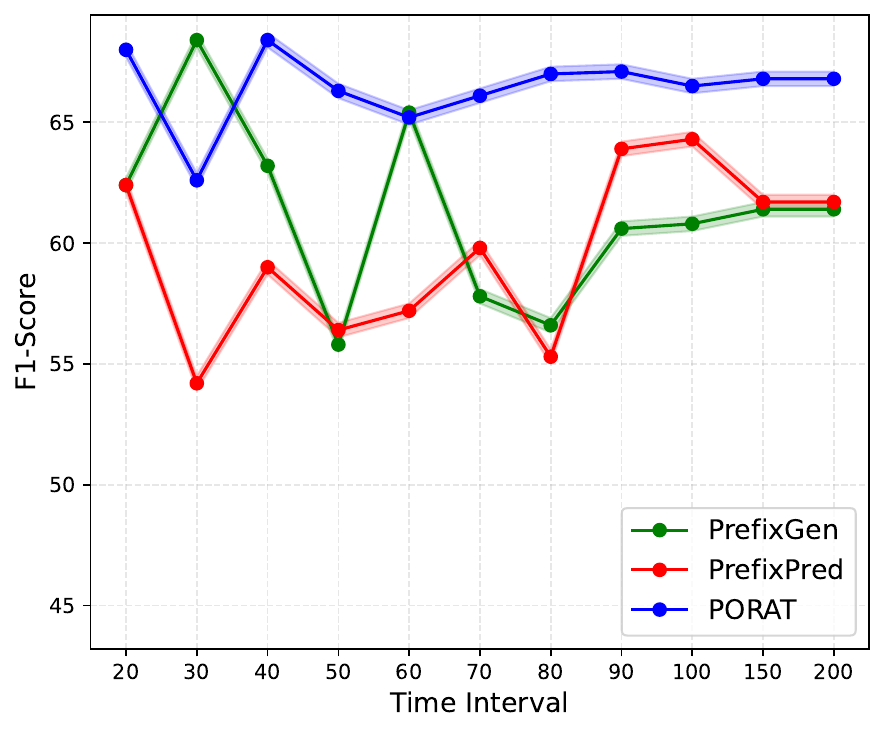}
}
\caption{Analysis of policy optimization timestep.}
\label{figure_ts}
\vspace{-16pt}
\end{figure}

\noindent \textbf{(2) Analysis of time interval for optimization.}  %
To gain an insight into the effects of selecting different interval of timestep $N$, we also conduct the analysis experiments varying $N$ from 1 to 10. 
As Fig.\ref{figure_ts}(a-b) show, we can observe that the impact of policy intervention time intervals is relatively minor. 
In contrast, applying policy interventions solely to either the generator or the predictor 
results in poorer model stability, whereas their joint presence leads to more stable performance.
This further strengthens the effectiveness of our proposed method and validates the feasibility of the theoretical framework. 
In addition, we also provide analysis with the longer interval from 20 to 200 in Fig.\ref{figure_ts}(c-d), which once again validates the previous conclusion.

\subsection{Visualization Analysis}
In Table \ref{tab:case}, we also visualize the rationales generated using recent model DR and our PORAT intuitively. 
We find that although both models provide correct predictions, DR instead correlates with other aspects. 
This indicates that DR has already been able to focus on explanations in the Aroma aspect. 
However, it fails to simultaneously address spurious correlations or other forms of suboptimal rationales well, while PORAT demonstrates superior capability.  
One possible reason is that DR only focuses on degeneration to employ a smaller lipschitz constant to capture semantically closer rationale candidates \cite{liu2023decoupled}. But this still limits its scalability in addressing other pattern problems.

\section{Conclusion and future work}\label{conclusion}
In this paper, we propose PORAT, a policy optimization-based data-centric self-explanation rationalization method. 
We first systematically revisit the cooperative game mechanism of rationalization in a novel game-theoretic perspective, and reveal the game-theoretic problem between two players in rationalization.
Then we theoretically analyze the causes of the game-theoretic problem between two players in rationalization and also prove the feasibility of the proposed method. 
Extensive experiments on nine widely used real-world datasets and two synthetic settings show that our proposed method significantly improves performance and outperforms several recently published SOTA methods. 
Furthermore, experiments on ablation studies, low-sparsity analysis, and language models analysis demonstrate the effectiveness and diversity of PORAT.
Moving forward, we plan to explore the feasibility of rationalizing predictions for large generative language models such as self-explanation foundation model, and further study other way to address the collapsed self-rationales in the field of rationalization.

\section*{Acknowledgment}

We would like to sincerely thank all the reviewers for their time and efforts in reviewing our paper. We also appreciate Mcauley et al. \cite{beer}, Lei et al. \cite{lei2016rnp} and Wang et al. \cite{hotel} for their dataset contributions. This work was supported by the National Natural Science Foundation of China (Nos.62376144, 62272285, 61906111), Natural Language Processing Innovation Team (Sanjin Talents) Project of Shanxi Province and the Science and Technology Cooperation and Exchange Special Project of Shanxi Province (No.202204041101016).

\bibliographystyle{ieeetr}
\bibliography{custom}

\newpage
\balance

\end{document}